\documentclass[10pt,twocolumn,letterpaper]{article}

\usepackage[pagenumbers]{cvpr} 
\usepackage{caption}
\usepackage{float}
\usepackage[table]{xcolor}
\usepackage{tabularx}

\usepackage{capt-of}
\usepackage{pifont}
\usepackage{booktabs}
\usepackage{multirow}
\usepackage{graphicx}
\usepackage{siunitx}
\newcommand{\TODO}[1]{}

\definecolor{cvprblue}{rgb}{0.21,0.49,0.74}
\definecolor{tabfirst}{rgb}{1, 0.7, 0.7}
\definecolor{tabsecond}{rgb}{1, 0.85, 0.7}
\definecolor{tabthird}{rgb}{1, 1, 0.7}
\usepackage[pagebackref,breaklinks,colorlinks,allcolors=cvprblue]{hyperref}

\usepackage{makecell}
\usepackage[accsupp]{axessibility}
\usepackage{algorithm}
\usepackage{algorithmicx}
\usepackage{algpseudocode}
\usepackage[export]{adjustbox}
\def\paperID{18063} 
\def\confName{CVPR}
\def\confYear{2026}

\title{PiLoT: Neural Pixel-to-3D Registration for UAV-based Ego and Target Geo-localization}

\definecolor{pilotpink}{HTML}{e64980}

\hypersetup{
    colorlinks=true,
    urlcolor=pilotpink
}
\author{
    Xiaoya Cheng$^{1}$ \quad Long Wang$^{2,3}$ \quad Yan Liu$^{4}$ \quad Xinyi Liu$^{1}$ \\
    Hanlin Tan$^{1}$ \quad Yu Liu$^{1}$ \quad Maojun Zhang$^{1}$ \quad Shen Yan$^{1\dagger}$ \\[4pt]
    {\small $^{1}$National University of Defense Technology \quad $^{2}$Zhejiang University} \\
    {\small $^{3}$Westlake University \quad $^{4}$Hangzhou Dianzi University} \\
    {\tt\small \{chengxy, liuxinyi24, hanlin\_tan, jasonyuliu, mjzhang, yanshen12\}@nudt.edu.cn} \\
    {\tt\small wanglong@westlake.edu.cn \quad 43038@hdu.edu.cn} \\[4pt]
    {\small \url{https://nudt-sawlab.github.io/PiLoT/}}
}

\begin{document}
\twocolumn[{
\renewcommand\twocolumn[1][]{#1}%
\maketitle
\vspace{-1cm}
\begin{center}
    \centering
    \captionsetup{type=figure}
    \includegraphics[width=\textwidth]{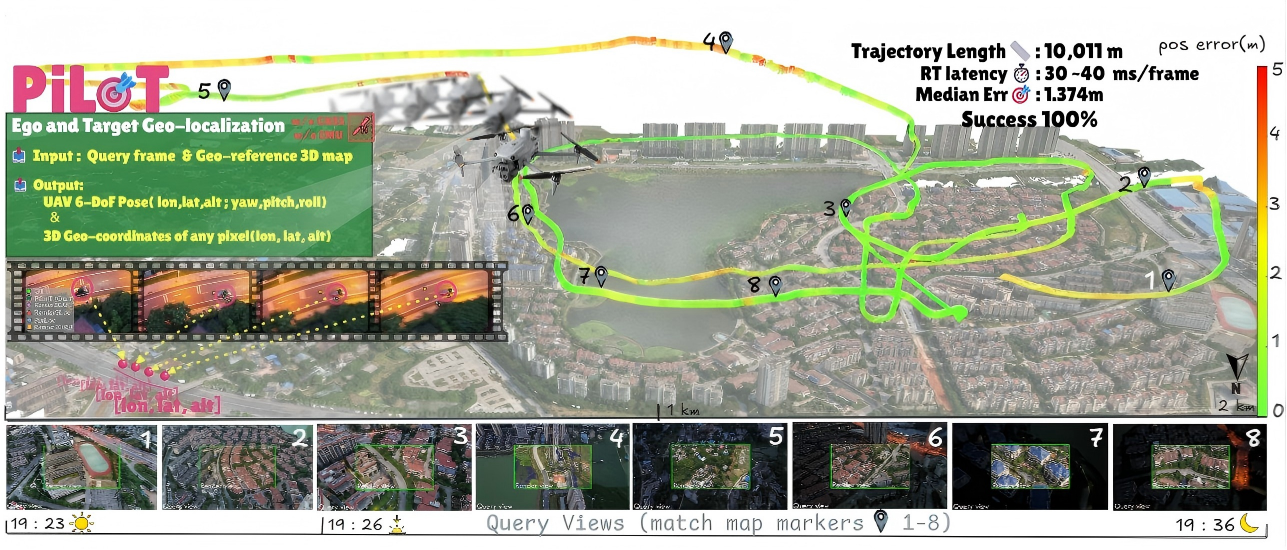}
    \vspace{-0.8cm}
    \captionof{figure}{\textbf{Overview of PiLoT.} Our system takes a live video frame and a geo-referenced 3D map as input, and outputs 1) the UAV 6-DoF pose, visualized by the tight alignment in the AR overlays (bottom row), and 2) the 3D geo-coordinates of any target pixel, as shown in the dynamic target tracking example (left, filmstrip view). PiLoT achieves \textcolor{blue}{drift-free}, \textcolor{red}{real-time}, and \textcolor{green}{long-term} ego and target geo-localization, demonstrated on a 10 km UAV trajectory with error color-coded (green: low, red: high). The system attains a median error of 1.37 m, a per-frame latency of 30 $\sim$ 40 ms, and 100\% success rate across day-to-night and cross-season variations without GNSS and IMU signals.}
    \label{fig:teaser}
\end{center}
}]

\begin{abstract}
We present PiLoT, a unified framework that tackles UAV-based ego and target geo-localization. 
Conventional approaches rely on decoupled pipelines that fuse GNSS and Visual-Inertial Odometry (VIO) for ego-pose estimation, and active sensors like laser rangefinders for target localization. However, these methods are susceptible to failure in GNSS-denied environments and incur substantial hardware costs and complexity.
PiLoT breaks this paradigm by directly registering live video stream against a geo-referenced 3D map. 
To achieve robust, accurate, and real-time performance, we introduce three key contributions: 1) a Dual-Thread Engine that decouples map rendering from core localization thread, ensuring both low latency while maintaining drift-free accuracy; 2) a large-scale synthetic dataset with precise geometric annotations (camera pose, depth maps). This dataset enables the training of a lightweight network that generalizes in a zero-shot manner from simulation to real data; and 3) a Joint Neural-Guided Stochastic-Gradient Optimizer (JNGO) that achieves robust convergence even under aggressive motion.
Evaluations on comprehensive public and newly collected benchmarks show that PiLoT outperforms state-of-the-art methods while running over 25 FPS on NVIDIA Jetson Orin platform.
\end{abstract}
    
\section{Introduction}
\label{sec:intro}
If a UAV with a single monocular camera could instantly localize itself in the world and geolocate everything it sees, it would unlock a new era of autonomy, enabling true-to-life digital twins, AR/VR applications, reliable navigation, and embodied AI for UAV. 
The mainstream approaches~\cite{qin2018vins,cao2022gvins} tackle the problem with a decoupled pipeline: localizing the UAV using visual-inertial odometry (VIO) fused with GNSS, and subsequently employing active sensors like laser rangefinders for target acquisition (e.g., DJI Matrice 4 Series). However, this paradigm suffers from two critical limitations: its reliance on GNSS makes it fragile in denied or degraded environments, and its laser-based targeting is costly, cumbersome, and restricted to a single point.

In this paper, we argue for a fundamental paradigm shift away from these decoupled, sensor-heavy systems. Our core idea is to reformulate UAV-based ego and target geo-localization as a unified pixel-to-3D registration problem. By continuously registering the live UAV video stream against a global 3D map (Google Earth for example), our system inherently recovers the UAV 6-DoF pose and geo-coordinates of any given pixel in the query image.
 
However, realizing such a system is non-trivial, as it requires resolving the fundamental trade-offs between accuracy, robustness and real-time performance, often termed the ``impossible triangle". 
This challenge is profoundly amplified by the demanding conditions of aerial deployment, specifically: 1) \textbf{Drift-Free Accuracy.} While VIO/SLAM methods~\cite{campos2021orb, qin2018vins, cao2022gvins, murai2025mast3r} could provide smooth localization, they inherently accumulate drift over long-duration flights. 2) \textbf{Environmental and Motion Robustness.} Localization is challenged by severe appearance variations (e.g., day-to-night, seasonal changes) between the captured video and the reference map. Concurrently, aggressive 6-DoF UAV motion can cause large inter-frame displacements, exceeding the basin of convergence of standard optimizers. 3) \textbf{Real-time Performance.} 
Recovering a globally consistent pose per frame is computationally intensive, creating a significant bottleneck for learning-based matchers~\cite{detone2018superpoint, sun2021loftr, sarlin2021back, panek2022meshloc, sarlin2019coarse, yan2023render} on resource-constrained onboard hardware (e.g., NVIDIA Jetson Orin).

To conquer these challenges, we propose PiLoT, a new paradigm for UAV-based ego and target localization. PiLoT is built upon three key technical contributions, each designed to address a side of the impossible triangle.
First, we introduce a \textbf{dual-thread framework} to decouple localization from map rendering. A \textit{Render Thread} generates a geo-referenced synthetic view on-the-fly, while a concurrent \textit{Localization Thread} registers the incoming video stream against this view in a feature space. This design ensures that every query frame is constrained by dynamically updated geo-anchors that follows the UAV's perspective, which is the key to achieve drift-free performance.

Second, we develop \textbf{a custom AirSim-Cesium-Unreal engine simulator}, and build \textbf{a new, million-scale synthetic dataset} by simulating UAV trajectories over vast photorealistic global terrains. By providing calibrated geometric supervision (metric depth, verified poses) across diverse conditions (weather, lighting), our dataset compels the network to learn features grounded in stable 3D geometry. Experiments confirm that these domain-invariant geometric cues enable our UAV-specific network to achieve zero-shot generalization on real-world data.

Third, we propose a \textbf{J}oint \textbf{N}eural-\textbf{G}uided Stochastic-Gradient \textbf{O}ptimizer (JNGO) that performs pixel-to-3D registration of the query frame. At its core, JNGO optimizes the camera pose by aligning the query frame with the projection of a reference 3D map in a shared feature space. 
To handle aggressive UAV motions, JNGO synergizes stochastic and gradient-based optimization for effective global exploration and local refinement. 
Specifically, it first generates a multitude of initial pose hypotheses. Each hypothesis is then refined in parallel via a gradient-based optimization that maximizes feature alignment, an operation we efficiently implement in CUDA. By repeating this process across multiple feature levels, JNGO achieves robust, real-time convergence even under extreme inter-frame displacements of up to 10 meters and 10 degrees of yaw.

To validate our approach and benefit the research community, we introduce a new, comprehensive benchmark suite for UAV-based ego and target geo-localization. Our benchmark comprises both challenging real-world sequences and large-scale synthetic data, specifically designed to test robustness against severe weather and lighting variations, aggressive motion, and long-term flights. Extensive experiments on public dataset and this benchmark demonstrate that our method substantially outperforms state-of-the-art vision-based approaches in both ego and target geo-localization. Furthermore, we validate its practical viability by deploying it on an NVIDIA Jetson Orin, where it achieves real-time performance with 25 FPS.

\section{Related Work}
\label{sec:related}
\noindent\textbf{UAV-based Ego localization.}
UAV-based ego localization aims to estimate the 6-DoF pose in a global coordinate. 
While SLAM and VIO methods~\cite{campos2021orb, murai2025mast3r} are robust for local state estimation, they are prone to drift in the absence of a global reference, limiting their universal applicability for UAVs.
To achieve drift-free localization, a dominant paradigm is to register the UAV's view against a geo-referenced map. 

Early methods used 2D satellite imagery~\cite{sarlin2023orienternet,vo2016localizing,lin2015learning}, yielding only 3-DoF pose (latitude, longitude, and yaw), and remain limited to 2D maps and simplified top-down assumptions.
To recover the full 6-DoF pose, recent works have turned to 3D maps~\cite{sarlin2019coarse,sarlin2021back, yan2023render,panek2022meshloc}. These methods typically initialize the pose via retrieval~\cite{hausler2021patch,berton2024meshvpr,arandjelovic2016netvlad} or sensor priors, and then refine it through either matching-based ~\cite{sun2021loftr, detone2018superpoint, sarlin2020superglue,zhou2021patch2pix, lindenberger2023lightglue} or direct alignment approaches ~\cite{sarlin2021back, niu2025hgsloc, kerbl20233d,yen2021inerf, huang2025sparse}. Matching-based methods are computationally too expensive and time-consuming for resource-constrained UAVs~\cite{sun2021loftr, lindenberger2023lightglue,detone2018superpoint, sarlin2020superglue,zhou2021patch2pix}. Alternatively, direct alignment methods, such as photometric optimization~\cite{yen2021inerf, kerbl20233d, niu2025hgsloc}, are highly sensitive to outdoor illumination, whereas feature-metric optimization provides better robustness but remains initialization-sensitive and generalizes poorly to aerial UAV views~\cite{sarlin2021back, huang2025sparse}. Our unified pixel-to-3D framework mitigates initialization sensitivity and closes the aerial domain gap via large-scale, map-grounded training.
\begin{table}[t]
\centering
\caption{
Comparison of UAV-oriented datasets. 
\emph{Legend:} \ding{51} = yes, \textcolor{gray}{\ding{55}} = no. 
\emph{Cond.} = illumination/weather changes; 
\emph{6-DoF} = availability of ground-truth 6-DoF pose; 
\emph{Depth} = availability of metric depth; 
\emph{Seq.} = contains continuous video sequences.
}
\vspace{-2mm}
\small
\setlength{\tabcolsep}{3pt}
\resizebox{\linewidth}{!}{
\begin{tabular}{lccccccccc}
\toprule
Dataset & Img & Region & Alt (m) & Pitch ($^{\circ}$) & Type  & Cond. & 6-DoF & Depth & Seq.\\
\midrule
University-1652~\cite{zheng2020university} & 37.8k & 1,652 & - & varied & Synthetic & \textcolor{gray}{\ding{55}}  & \textcolor{gray}{\ding{55}} & \textcolor{gray}{\ding{55}}& \textcolor{gray}{\ding{55}} \\
SUES-200~\cite{zhu2023sues} & 24.1k & 200 & 150--300 & varied & Real & \textcolor{gray}{\ding{55}} & \textcolor{gray}{\ding{55}} &\textcolor{gray}{\ding{55}} & \ding{51}\\
DenseUAV~\cite{dai2023vision} & 27k & 14  & 80--100  & fixed & Real & \ding{51} & \textcolor{gray}{\ding{55}}& \textcolor{gray}{\ding{55}} & \textcolor{gray}{\ding{55}} \\
UAV-VisLoc~\cite{xu2024uav} & 6.7k & 11 & 400--2000 & fixed & Real & \ding{51} & \textcolor{gray}{\ding{55}} & \textcolor{gray}{\ding{55}}& \ding{51} \\
UAVD4L~\cite{wu2024uavd4l} & 6.8k & 1 & 50--300 & varied & Real & \textcolor{gray}{\ding{55}} & \ding{51}& \ding{51}& \textcolor{gray}{\ding{55}} \\
GAME4Loc~\cite{ji2025game4loc} & 33.7k & 1 & 80--650 & varied & Synthetic & \textcolor{gray}{\ding{55}} & \ding{51}& \textcolor{gray}{\ding{55}} & \textcolor{gray}{\ding{55}}\\
MatrixCity~\cite{li2023matrixcity} & 67k & 2 & 100--450 & fixed & Synthetic & \ding{51}& \ding{51}& \ding{51} & \ding{51} \\
OrthoLoC~\cite{dhaouadi2025ortholoc} & 16.4k & 47 & 50--300 & varied & Real & \textcolor{gray}{\ding{55}} & \ding{51}& \ding{51}& \textcolor{gray}{\ding{55}} \\
UAVScenes~\cite{wang2025uavscenes} & 120k & 4 & 80--130 & fixed & Real & \textcolor{gray}{\ding{55}} & \ding{51}& \ding{51}& \ding{51} \\
\midrule
\textbf{Ours} & \textbf{1M+} & \textbf{378} & \textbf{$\sim$800} & \textbf{varied} & \textbf{Synthetic}  & \textbf{\ding{51}} & \textbf{\ding{51}} & \textbf{\ding{51}} & \textbf{\ding{51}}\\
\bottomrule
\end{tabular}
}
\vspace{-3mm}
\label{tab:uav-dataset-type}
\end{table}

\noindent\textbf{UAV-based Target Geo-localization.}
UAV-based target geo-localization determines the 3D world coordinates of a target from an image. A primary approach relies on geometric principles, combining camera models with UAV poses to infer target coordinates~\cite{zhou2025moving, cai2022review}. This method is highly sensitive to the quality of the pose estimate. To improve accuracy, 3D map-based geolocation~\cite{wu2024uavd4l} has become a mainstream approach, using techniques like rendering-based matching and DSM projection to obtain precise target coordinates. While accurate, these methods are often bottlenecked by the need for computationally expensive rendering-based matching or tight coupling with a non-real-time ego-localization pipeline. This prevents them from achieving the millisecond-level response required for dynamic targeting applications, a gap our work aims to fill.

\noindent\textbf{UAV-Oriented Visual Localization Datasets.}
Although recent visual localization methods have made significant progress, they remain largely tailored to ground-level tasks. Models trained on such data often fail in aerial domains, highlighting the need for UAV specific datasets.
As detailed in \cref{tab:uav-dataset-type}, while many ~\cite{zheng2020university, zhu2023sues,dai2023vision,xu2024uav,wu2024uavd4l,ji2025game4loc,li2023matrixcity,wang2025uavscenes} offer photorealism, they often lack the scale and complete geometric ground truth (e.g., 6-DoF poses, metric depth) needed for robust training.
Even recent efforts with geometric data~\cite{li2023matrixcity, wang2025uavscenes} are limited in city scale and viewpoint variety, making them insufficient for sequential geometric supervision.
To address the lack of structured UAV training data, we build a fully automated AirSim–Cesium–Unreal pipeline that converts large-scale geospatial data into geo-aligned imagery with precise 6-DoF poses and depth maps. 

\section{Method}
\label{sec:method}
\subsection{Overview}
\label{sec:overview}

Given a geo-referenced 3D map $\mathcal{M}$, a monocular video stream $\{I_i^q\}$ with known intrinsics $\{\mathbf{K}_i\}$, and a single pose prior for the first frame~($\tilde{\mathbf{T}}_{init}$), we address the problem of UAV-based ego localization and target geo-localization without aid from GNSS and IMU. Specifically, our goals are twofold: 1) estimate $\hat{\mathbf{T}}_i$ for every query frame, and 2) enable precise pixel-to-geo projection that maps any query pixel $\mathbf{u}=(u,v)^\top$ on frame $\{I_i^q\}$ to its real-world coordinates $(\mathrm{lon},\mathrm{lat},\mathrm{alt})$.

\begin{figure}[t!]
  \centering
  \includegraphics[width=\linewidth]{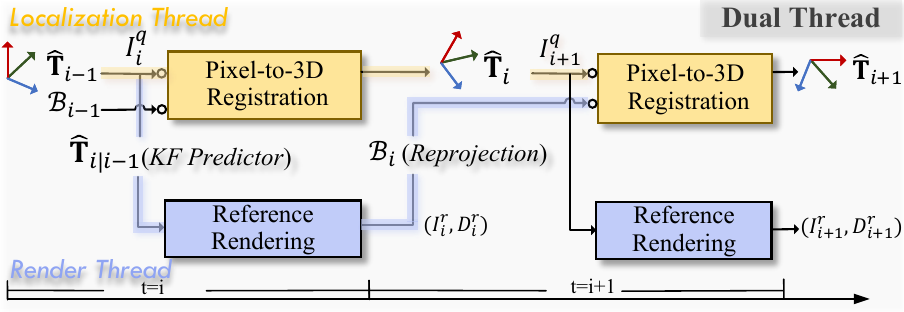}
  \caption{\textbf{PiLoT's Dual-Thread Framework}. We decouple rendering from localization into two parallel threads. A Render Thread generates synthetic views, while a concurrent Localization Thread registers the live frame against them to compute the pose, ensuring high-frequency accuracy.}
  \label{fig:method_framework}
  \vspace{-3mm}
\end{figure}

\begin{figure*}[t!]
  \centering
  \includegraphics[width=\linewidth]{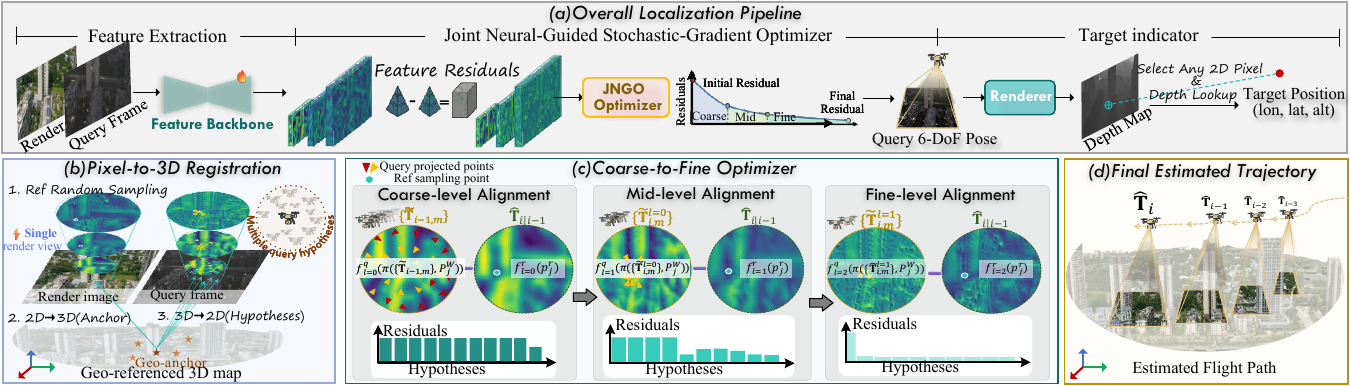}
  \caption{\textbf{Overview of the PiLoT framework and localization pipeline.} 
  \textbf{(a)} The overall pipeline inputs a query frame and outputs the UAV's 6-DoF ego-pose along with the target's 3-DoF geo-location. 
  \textbf{(b)} A highly efficient one-to-many paradigm matches multiple query hypotheses against a \textit{single} rendered reference view via feature alignment.
  \textbf{(c)} Our coarse-to-fine optimizer iteratively narrows the search space to converge on the optimal 6-DoF pose. 
  \textbf{(d)} The final estimated trajectory demonstrates robust and drift-free sequential localization.}
  \label{fig:method_pipeline}
  \vspace{-3mm}
\end{figure*}

\subsection{PiLoT's Dual-Thread Framework}
\label{sec:engine}
For sequential video localization, a naive strategy is to render a reference view from the last frame and perform pose refinement for the current frame. This linear dependency creates an inherent temporal bottleneck, where the localization engine is forced to stall until the rendering task completes. Instead of this conventional linear pipeline, we propose a decoupled dual-thread architecture that synchronizes map rendering and pose optimization in parallel (\cref{fig:method_framework}). 

Within this framework, a common strategy to handle rapid motion is to render multiple views around the last estimated pose, thereby expanding the search region. Our system instead strategically renders a single reference anchor and leverages a one-to-many strategy. This approach refines a swarm of pose hypotheses against the shared rendering, achieving a wide search range without multiple reference viewpoints. The two threads coordinate this process as follows:

\noindent\textbf{Rendering thread.}
This thread runs to provide a geo-referenced view for localization.
As illustrated in \cref{fig:method_framework} (bottom), the rendering thread first predicts a reference pose $\hat{\mathbf{T}}_{i|i-1}$ from the last estimate $\hat{\mathbf{T}}_{i-1}$ using a constant-velocity Kalman filter (KF). From this predicted pose, we then render a new reference view ($I_{i}^r, D_{i}^r$) and back-project $N$ of its depth-valid pixels into the world frame to form a set of 3D geo-anchors:
\begin{equation} 
  \mathbf{P}_{i,j}^{W} = \hat{\mathbf{T}}_{i|i-1} \left( D_{i}^{r}(\mathbf{p}^r_{i,j}) \cdot \mathbf{K}^{-1}\mathbf{p}^r_{i,j} \right) \label{eq:back_project} 
\end{equation} 

We finally package a reference bundle
\begin{equation}
\mathcal{B}_i \;:=\; \big(I_i^{r},\, \mathbf{\hat{T}}_{i|i-1}, \{\mathbf{P}_{i,j}^{W}\}_{j=1}^{N}\big)
\end{equation}
and pass it to the localization thread.

\noindent\textbf{Localization thread.} 
As depicted in \cref{fig:method_framework} (top), our Pixel-to-3D Registration pipeline executes for each a new query frame $I_{i+1}^q$. It begins by extracting multi-scale features and uncertainty maps from both the query and the reference view $I_{i}^{r}$ using a lightweight extractor (\cref{sec:extractor}). Anchored by the reference bundle $\mathcal{B}_i$, our JNGO optimizer (\cref{sec:jngo_optimizer}) then performs a global exploration with local exploitation to conduct a wide-area search and find the globally consistent pose estimate $\mathbf{\hat{T}}_{i+1}$. This new pose is subsequently passed back to the rendering thread to prepare the reference bundle $\mathcal{B}_{i+1}$ for the next cycle.

For the sake of readability, we will omit the frame index i from our notations in the following sections when the context is clear, for example, using ${P^W_j}$ instead of ${P^W_{i,j}}$.

\begin{figure*}[t!]
  \centering
  \includegraphics[width=\linewidth]{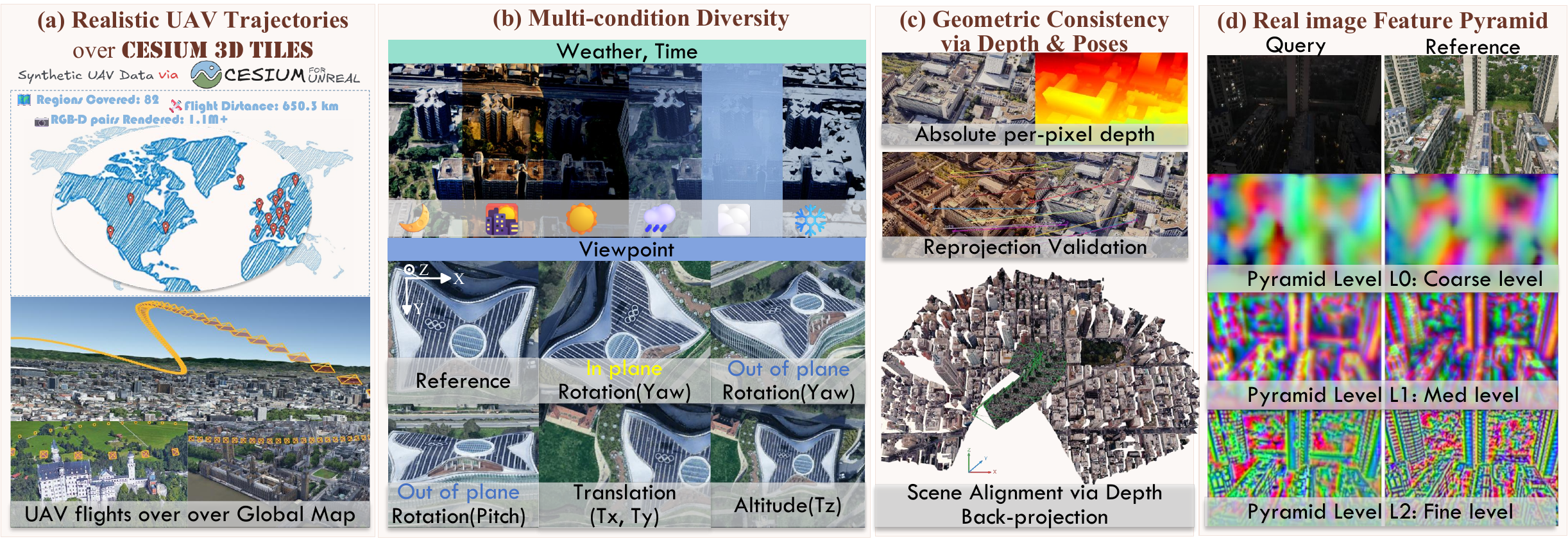}
  \caption{\textbf{Overview of our synthetic data generation and its resulting zero-shot sim-to-real performance.}
  From left to right:
  (a) realistic UAV trajectories rendered over geo-referenced 3D tiles in Cesium for Unreal;
  (b) multi-condition diversity across weather/time and viewpoint (in-plane yaw, out-of-plane pitch/yaw, planar translation $T_x,T_y$, altitude $T_z$);
  (c) geometric consistency: we export \emph{absolute per-pixel depth} and validate by reprojection; 
  (d) \emph{our} three-level feature pyramid on query (real) vs. reference (synthetic) images.
  }
  \label{fig:method_dataset}
  \vspace{-3mm}
\end{figure*}
\subsection{Pixel-to-3D Registration}
\label{sec:network}

\subsubsection{UAV-specific Feature Extraction}
\label{sec:extractor}
\noindent\textbf{Lightweight neural network.}
We seek UAV-specific features that remain discriminative under large viewpoint and illumination changes while running at edge speed. We adopt an off-the-shelf MobileOne-S0 encoder (depth=3, ImageNet-initialized) with a compact U-Net decoder, shared by the query $I^q$ and reference $I^{r}$ branches. Given an $H \times W$ RGB image, it outputs a three-level pyramid at $1/4$ (coarse), $1/2$ (mid), and $1$ (fine) resolution with a compact channel width $C{=}32$, yielding the query features and uncertainties $\{(\mathbf{f}_\ell^{q},\,w_\ell^{q})\}_{\ell=0}^{2}$ and the reference counterparts $\{(\mathbf{f}_\ell^{r},\,w_\ell^{r})\}_{\ell=0}^{2}$. For method details, please refer to \cref{sec:sup_network_arch} in the Appendix.

\noindent\textbf{Training with a large-scale dataset.}
\label{sec:dataset}
Training this neural network hinges on large-scale datasets with dense depth and precise camera poses for geometric supervision.
However, existing UAV datasets~\cite{zheng2020university, zhu2023sues, dai2023vision, ji2025game4loc} are deficient in such labels and scale. Even recent efforts with geometric data~\cite{ li2023matrixcity, wang2025uavscenes} are limited in a few city models and viewpoint variety, making them insufficient for sequential geometric supervision.
To bridge this critical data gap, we introduce a new large-scale synthetic dataset specifically designed to support geometry-aware learning.

We develop a fully automated simulator based on the \textbf{AirSim-Cesium-Unreal Engine pipeline}. Using this powerful tool, we generate \textbf{a new, million-scale synthetic dataset} by simulating flights over vast, photorealistic global terrains. As illustrated in \cref{fig:method_dataset}(a-c), our dataset provides RGB and pixel-wise depth images captured along realistic UAV trajectories under diverse visual conditions (e.g., scenes, weather, lighting). Crucially, we provide precise and geometrically-consistent ground truth, including absolute camera poses, all rigorously validated through reprojection. Please see \cref{sec:supp_data_generation} in the Appendix for details on our dataset generation and statistics.

By providing accurate geometric supervision across diverse visual conditions, our large-scale dataset compels the network to learn features grounded in the underlying 3D structure. This makes the learned representations inherently robust to photometric variations and is the key to training a lightweight UAV-specific network that achieves zero-shot generalization to real-world data, as illustrated in \cref{fig:method_dataset}(d).

\noindent\textbf{Supervision.}
\label{sec:training}
We train our network using a direct alignment approach, jointly optimizing the feature extractor and the subsequent iterative pose refinement process end-to-end. The core of our training is a geometric loss that minimizes the reprojection error between the ground-truth 2D projections $\mathbf{p}_j^q$ of the geo-anchors and their estimated projections $\tilde{\mathbf{p}}_j^q$ based on the pose estimate:
\begin{equation}
\mathcal{L} = \sum_{j} \rho_B \left( \left\| \mathbf{p}_j^q - \tilde{\mathbf{p}}_j^q \right\|_2^2 \right)
\end{equation}
\vspace{-1mm}

where $\rho_B(\cdot)$ is Barron's robust loss function ~\cite{barron2019general}.

\subsubsection{Joint Neural-Guided Stochastic-Gradient Optimizer}
\label{sec:jngo_optimizer}
Aggressive UAV motion often induces large inter-frame displacements, posing a significant challenge for traditional gradient-based optimizers that are prone to local minima. To address this, we introduce the JNGO, which navigates the challenging, non-convex optimization landscape by synergizing a global exploration with local exploitation.

\begin{figure}[t!]
    \centering
    \includegraphics[width=\columnwidth]{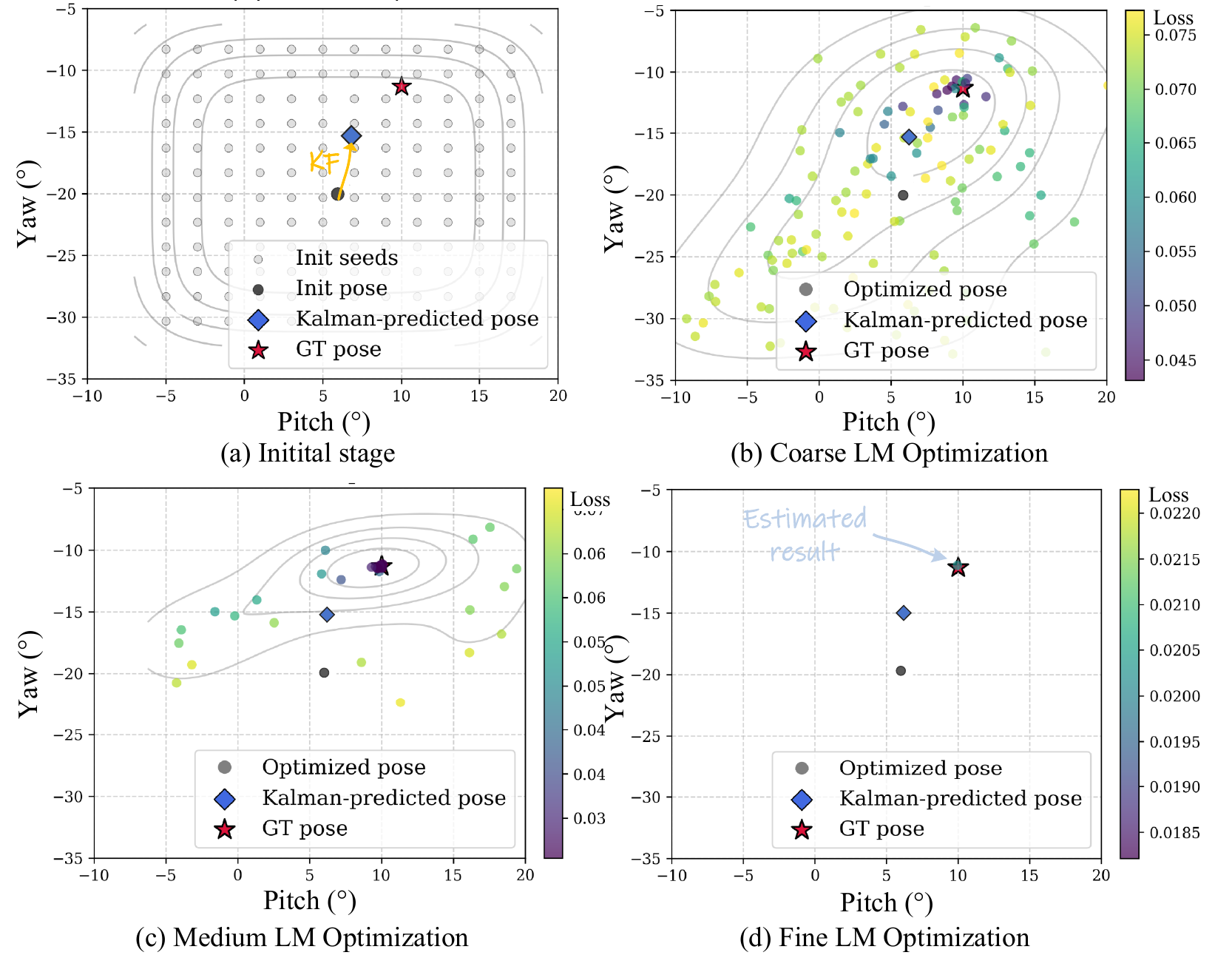}
    \caption{\textbf{Rotation-aware sampling and coarse-to-fine optimization.}
    The figure visualizes the pose convergence process in the pitch/yaw space: it synergizes wide-area Rotation-Aware Sampling (a) with parallel, coarse-to-fine refinement (b-d) to ensure robust convergence under aggressive motion.
    }
    \label{fig:KaMH}
    \vspace{-1mm}
\end{figure}

\noindent\textbf{Rotation-Aware Hypothesis Generation.}
Based on the observation that apparent pixel displacement in UAV imagery is far more sensitive to rotations than to translations, we design a Rotation-Aware Sampling strategy to generate the hypotheses $\tilde{\mathbf{T}}_{m}$ by adaptively enlarges the search range along motion-sensitive axes, pitch and yaw. As shown in ~\cref{fig:KaMH} (a), centered at the hypotheses on the previous frame, rotational perturbations are sampled uniformly from an anisotropic bounding box $\mathcal{B}_r$ that allocates greater range to pitch and yaw:
\begin{equation}
\begin{gathered}
\mathcal{B}_r =
[-\alpha_{\text{pitch}},\alpha_{\text{pitch}}]\times
[-\alpha_{\text{yaw}},\alpha_{\text{yaw}}]
\end{gathered}
\end{equation}
and minor translational perturbations are drawn from a Gaussian distribution: 
\begin{equation}
\begin{aligned}
\delta\mathbf{t}_m &\sim \mathcal{N}\!\big(\boldsymbol{\mu}_t,\boldsymbol{\Sigma}_t\big),\\[-2pt]
\delta\boldsymbol{\phi}_m &\sim \mathcal{U}(\mathcal{B}_r),\ \text{for } m=1,\dots,M.
\end{aligned}
\end{equation}
where \(\boldsymbol{\mu}_t,\boldsymbol{\Sigma}_t\) are inferred from the Kalman predictor. 

\noindent\textbf{Neural-Guided Parallel Refinement.}
Each hypothesis ${\tilde{\mathbf{T}}_{m}}$ is then refined in parallel using a coarse-to-fine Levenberg–Marquardt (LM) optimizer over the geo-anchors, with a small number of iterations per pyramid level, as shown in \cref{fig:KaMH}(b-d). At each pyramid level $\ell$, we refine the pose hypothesis $\tilde{\mathbf{T}}_{m}$ by minimizing the feature-based photometric cost $\mathcal{C}_{\text{photo}}^{(m,\ell)}$, which measures the residual $r_{j,\ell}^{(m)}$ between bilinearly sampled query features and reference features at a set of geo-anchors ${P^W_j}$:
\begin{equation}
\mathbf{r}^{(\ell)}_{j,m} = \mathbf{f}^q_{\ell}\left( \pi\left( \mathbf{K}_{\ell} , \tilde{\mathbf{T}}_{m}^{-1} , \mathbf{P}_j^W \right) \right) - \mathbf{f}^r_{\ell}(p^r_j)
\label{eq:residual}
\end{equation}
where $\pi(\cdot)$ is the pinhole model.
These per-anchor residuals are then aggregated into a cost function:

\begin{equation}
\mathcal{C}_{\text{photo}}^{(m, \ell)} = \sum_j \rho\left( w_\ell(j) \cdot \| \mathbf{r}^{(\ell)}_{j,m} \|^2_2 \right)
\label{eq:energy}
\end{equation}
where $\rho(\cdot)$ is the Huber robust loss~\cite{huber1992robust}, and $w_\ell(j)$ is a joint uncertainty score derived from $w_\ell^{q}$ and $w_\ell^{r}$ at the corresponding pixel locations.

This update is performed iteratively for a total of $K$ steps. At each iteration $k$, the pose is updated as:
\begin{align}
\left(\mathbf{J}^\top \mathbf{W} \mathbf{J} + \lambda \mathbf{I}\right) \Delta \boldsymbol{\xi} &= -\mathbf{J}^\top \mathbf{W} \mathbf{r}, \\
\tilde{\mathbf{T}}^{(k+1)}_{m} &= \exp\left( \Delta \boldsymbol{\xi} \right) \cdot \tilde{\mathbf{T}}^{(k)}_{m}.
\end{align}
where $\Delta \boldsymbol{\xi} \in \mathbb{R}^6$ is the Lie algebra increment, $\mathbf{J}$ is the Jacobian of residuals $\mathbf{r}$ with respect to $\boldsymbol{\xi}$, and $\mathbf{W}$ is a diagonal matrix of uncertainty scores. Here, $\mathbf r,\mathbf J,\mathbf W$ are built at step $(m,\ell,k)$. The exponential map $\exp(\cdot)$ denotes the SE$(3)$ pose update. Further implementation details are provided in \cref{sec:sup_lm_formulation} of the Appendix.
After $K$ iterations, the initial pose hypotheses are refined to $\{\tilde{\mathbf{T}}^{\prime}_{m}\}$.

\noindent\textbf{Motion-Constrained Hypothesis Selection.}
To robustly select the best pose from multiple hypotheses, we additionally leverage the physics-based motion prior that favors poses registering with the predicted trajectory.
We denote the KF predicted pose for the current frame as $\hat{\mathbf{T}}_{\text{pred}}$. 
Each hypothesis is scored by a total cost, which combines its final feature-based photometric cost $\mathcal{C}_{\text{photo}}^{(m,\ell=2)}$, with a motion regularization term:
\begin{equation}
\mathcal{C}_{\text{total}}^{(m)} = \mathcal{C}_{\text{photo}}^{(m,\ell=2)} + \lambda\,\|\log(\hat{\mathbf{T}}_{\text{pred}}^{-1} \tilde{\mathbf{T}}^{'}_{m})^\vee\|_2^2.
\label{eq:total_loss_combined}
\end{equation}
Here, $\mathcal{C}_{\text{motion}}^{(m)}$ computes the squared geodesic distance between the $m$-th hypothesis pose $\tilde{\mathbf{T}}^{'}_{m}$ and the KF-predicted pose $\hat{\mathbf{T}}_{\text{pred}}$. This distance is formulated in the Lie algebra $\mathrm{se}(3)$ by mapping the relative transformation to its 6D twist vector representation using the $\log(\cdot)^\vee$ operation. The hyperparameter $\lambda$ balances the data and motion terms.
The final pose $\hat{\mathbf{T}}$ is determined by selecting the most reliable hypothesis, the one with the minimum total loss:
\begin{equation}
m^* = \operatorname*{argmin}_{m} \mathcal{C}_{\text{total}}^{(m)}, \quad \hat{\mathbf{T}} = \tilde{\mathbf{T}}^{'}_{m^*}.
\label{eq:wta_selection}
\end{equation}
Given $\hat{\mathbf{T}}$, any query pixel can be mapped to geographic coordinates by casting a camera ray into $\mathcal{M}$ using the depth rendered from the estimated pose $\hat{\mathbf{T}}$.

\section{Experiments}
\label{sec:experiments}

\subsection{Experimental Setup}
\noindent\textbf{Implementation Details.}
We train our model using our proposed large-scale synthetic dataset (as detailed in \cref{sec:dataset}). This dataset provides training data in the form of reference-query pairs, sampled as consecutive frames along each generated UAV trajectory. We perturb query poses with random noise ($5 \sim 15$ m translation, $5 \sim 15^\circ$ rotation) to simulate initialization uncertainty. Augmentations combine high-frequency (Fourier) noise ~\cite{chattopadhyay2023pasta} with photometric jitter (blur, contrast, Gaussian noise, brightness). Supervision is provided by a geometric reprojection loss on $N=500$ reference geo-anchors. We train for 30 epochs using Adam (lr=$10^{-3}$) on 8 RTX 4090 GPUs.

At test time, we sample $M{=}144$ pose hypotheses by pitch/yaw in [$-11^\circ,11^\circ$] at $2^\circ$ steps and add translation noise which obeys ~$\mathcal{N}(0,1)$ m. Each hypotheses is optimized using 500 sampling geo-anchors.
We simulate a realistic initialization by assuming a coarse pose prior for the first frame with random noise of up to $10$ m in translation and $10^\circ$ in rotation, which is sourced from the ground truth of the first frame (for synthetic data) or coarse GNSS/IMU (for real-world data).

\noindent\textbf{Baselines.}
We benchmark against methods in the context of map-based, scene-generalizable localization. 
The first category consists of hybrid methods that combine frame-to-frame Visual Odometry (VO) with absolute pose corrections. We use ORB-SLAM3 (sparse feature-based)~\cite{campos2021orb} and RAFT (optical flow-based)~\cite{teed2020raft} for relative tracking, periodically corrected at 1Hz by a Render-and-Compare module~\cite{yan2023render}. This module renders a reference view, finds correspondences with a LoFTR~\cite{sun2021loftr} matcher, and computes the absolute pose via PnP. We term these methods as Render2ORB and Render2RAFT.

The second category performs per-frame absolute localization, where each query frame is localized independently against a newly rendered reference view. It includes PixLoc~\cite{sarlin2021back}, which solves for the pose via direct alignment of dense feature maps, and Render2Loc~\cite{yan2023render}, which instead employs a feature matching and PnP pipeline. For Render2Loc, we report results using four prominent matchers: LoFTR~\cite{sun2021loftr}, EfficientLoFTR (ELoFTR)~\cite{wang2024efficient}, and two recent SoTA matchers in aerial vision, Aerial-MASt3R~\cite{vuong2025aerialmegadepth} and RoMaV2~\cite{edstedt2025roma}.

\label{subsec:synthetic_eval}
\begin{figure}[t]
\centering
\includegraphics[width=\linewidth]{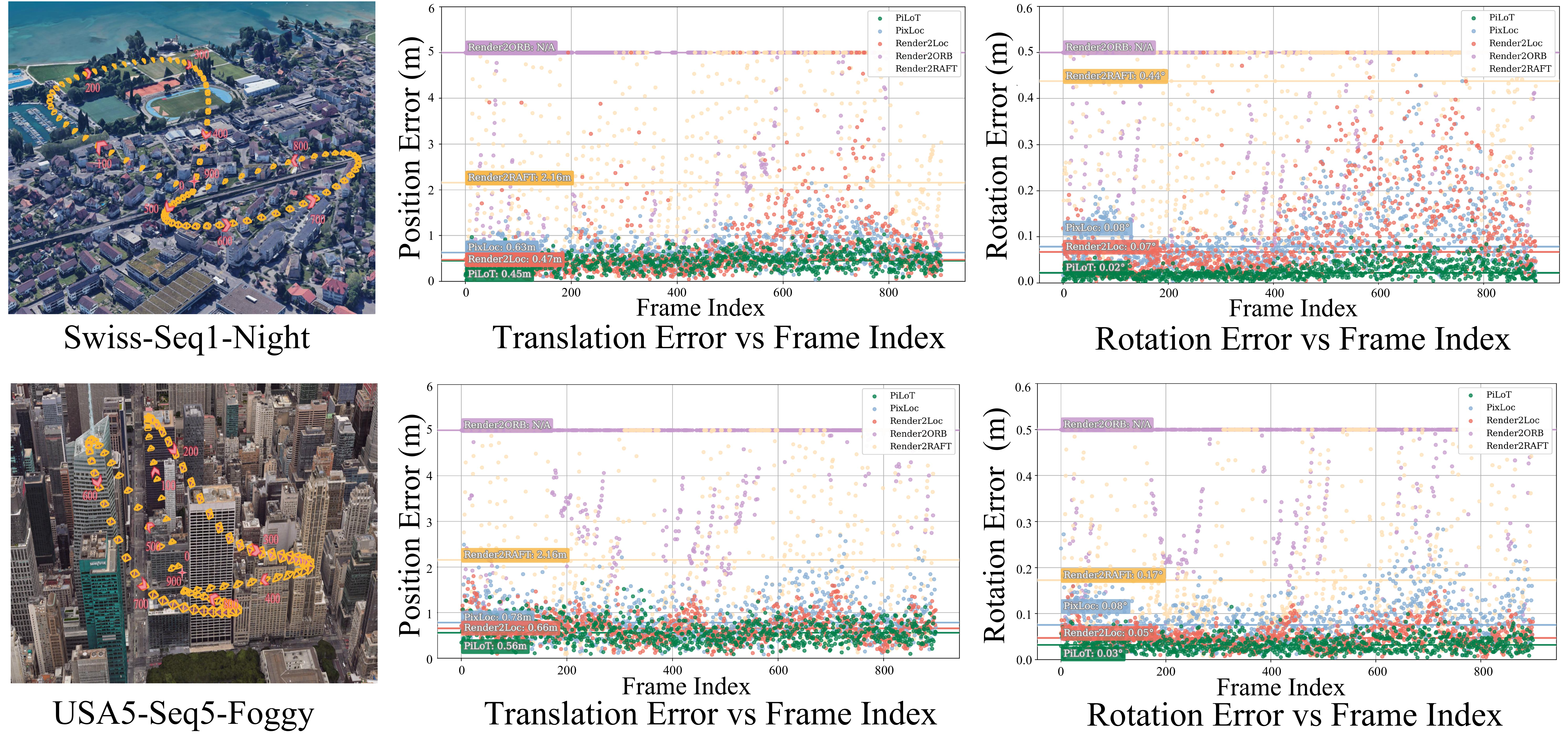} 
\vspace{-5mm}
\caption{
    \textbf{Per-frame stability on synthetic trajectories.} Two trajectories with per-frame translation and rotation errors, showing that PiLoT (green) sustains a lower error distribution with fewer catastrophic outliers.
}
\label{fig:exp_google_error}
\vspace{-4mm}  
\end{figure}
\definecolor{ourcolor}{RGB}{220, 242, 220}
\definecolor{secondbest}{RGB}{235, 250, 235}
\definecolor{thirdbest}{RGB}{250, 252, 240}

\newcommand{\best}[1]{\textbf{#1}}
\newcommand{\bestbox}{\cellcolor{ourcolor}}
\newcommand{\sndbox}{\cellcolor{secondbest}}
\newcommand{\trdbox}{\cellcolor{thirdbest}}

\begin{table*}[t]
\centering
\sisetup{
    detect-weight,
    mode=text,
    table-format=1.2,
    output-decimal-marker={.},
}
\caption{
    \textbf{Comprehensive localization performance on synthetic and real-world datasets.}
    We compare PiLoT against baselines across three diverse datasets, all using a shared map and 512\,px inputs. Median errors are reported in meters (m) and degrees ($^\circ$).
}
\vspace{-2mm}
\resizebox{\textwidth}{!}{%
\begin{tabular}{@{} l S[table-format=2.1] S[table-format=1.2] S[table-format=1.2] c S[table-format=3.1]
    S[table-format=1.2] S[table-format=1.2] c S[table-format=3.1]
    S[table-format=1.2] S[table-format=1.2] c S[table-format=3.1] @{}}
\toprule
& & \multicolumn{4}{c}{\textbf{SynthCity-6} (Synthetic)} & \multicolumn{4}{c}{\textbf{UAVScenes} (Real)} & \multicolumn{4}{c}{\textbf{UAVD4L-2yr} (Real)} \\
\cmidrule(lr){3-6} \cmidrule(lr){7-10} \cmidrule(l){11-14}
\textbf{Method} & {\textbf{FPS}$\uparrow$} & {Med m$\downarrow$} & {Med $^\circ$$\downarrow$} & {\textbf{R@1/3/5}~(m,${}^\circ$)$\uparrow$} & {Comp.$\uparrow$} & {Med m$\downarrow$} & {Med $^\circ$$\downarrow$} & {\textbf{R@1/3/5}~(m,${}^\circ$)$\uparrow$} & {Comp.$\uparrow$} & {Med m$\downarrow$} & {Med $^\circ$$\downarrow$} & {\textbf{R@1/3/5}~(m,${}^\circ$)$\uparrow$} & {Comp.$\uparrow$} \\
\midrule
\multicolumn{14}{@{}l}{\textit{Hybrid Methods}} \\
\addlinespace[0.2em]
Render2ORB        & \sndbox 20.0 & 4.37 & 0.21 & 18.0 / 30.2 / 55.3 & 38.4 & 11.36 & 6.62 & 5.7 / 63.5 / 78.4 & 72.3 & 3.53 & 4.34 & 29.1 / 70.9 / 89.1 & 66.7 \\
Render2RAFT$^{\dagger}$ & \trdbox 10.0 & 5.21 & 0.62 & 8.4 / 32.5 / 48.9  & 96.2 & 10.54 & 4.21 & 4.2 / 45.8 / 62.3 & 88.5 & 2.68 & 2.51 & 22.4 / 62.6 / 77.4 & 92.3 \\
\midrule
\multicolumn{14}{@{}l}{\textit{Absolute Localizers}} \\
\addlinespace[0.2em]
PixLoc$^{\dagger}$        & 5.0  & 0.55 & 0.06 & 65.4 / 92.3 / 98.2 & 96.5 & 2.85 & 0.94 & 12.8 / 94.6 / 97.5 & \trdbox 98.4 & 2.47 & 2.34 & 31.5 / 68.9 / 87.4 & 86.5 \\
Render2Loc (LoFTR)$^{\dagger}$   & 2.0  & \trdbox 0.49 & \sndbox 0.04 & \trdbox 76.5 / 98.1 / 99.2 & \bestbox 100.0 & \trdbox 1.62 & \sndbox 0.52 & \sndbox 23.2 / 95.8 / \best{98.9} & \bestbox 100.0 & \trdbox 1.08 & \sndbox 0.95 & \sndbox {44.2 / 90.5 / \best{98.2}} & \bestbox 100.0 \\
Render2Loc (ELoFTR)$^{\dagger}$  & \sndbox 20.0 & 0.51 & \sndbox 0.04 & 75.1 / 97.8 / 99.0 & \bestbox 100.0 & 1.84 & 0.65 & 20.6 / 95.2 / 98.4 & \bestbox 100.0 & 1.15 & 1.08 & \trdbox {44.0 / 86.8 / 97.7} & 98.7 \\
Render2Loc (Aerial-MASt3R)$^{\star}$ & 0.4  & 0.52 & 0.05 & 69.8 / 97.5 / 99.1 & \bestbox 100.0 & 1.85 & 0.72 & 21.2 / 94.8 / 97.9 & \bestbox 100.0 & 1.35 & 0.99 & 36.5 / 82.4 / 93.1 & \bestbox 100.0 \\
Render2Loc (RoMaV2)$^{\star}$  & 0.8  & \sndbox 0.47 & \sndbox 0.04 & \sndbox {77.2 / 98.8 / 99.6} & \bestbox 100.0 & \sndbox 1.42 & \trdbox 0.58 & \trdbox {22.1 / 95.5 / 98.2} & \bestbox 100.0 & \sndbox 1.05 & \trdbox 0.97 & 43.2 / \best{93.6} / 98.1 & \bestbox 100.0 \\
\midrule
\rowcolor{ourcolor}
\textbf{PiLoT (Ours)} & \best{28.0} & \best{0.46} & \best{0.03} & \best{80.4 / 99.9 / 99.9} & \best{100.0} & \best{1.27} & \best{0.47} & \best{25.5 / 95.9 }/ 98.4 & \best{100.0} & \best{0.92} & \best{0.89} & \best{45.8} / 92.1 / 97.9 & \best{100.0} \\
\bottomrule
\end{tabular}%
}
\vspace{-3pt}
\leftline{\scriptsize \quad $^{\dagger}$ Fine-tuned on our synthetic benchmark. $^{\star}$ SoTA matchers in aerial vision.}
\label{tab:overall_comparison}
\vspace{-2mm}  
\end{table*}

\subsection{UAV-based Ego Localization}
\noindent\textbf{Datasets.} 
We evaluate our method on three diverse benchmarks. The synthetic \textbf{SynthCity-6} dataset, generated with Cesium for Unreal, covers 60 UAV trajectories ($54\text{k}$ frames) over six $2\,\mathrm{km}\times2\,\mathrm{km}$ regions under various weather and lighting conditions.

We also test on the public \textbf{UAVScenes}~\cite{wang2025uavscenes} benchmark ($51.6\text{k}$ frames from 01 runs across AMtown, AMvalley, HKairport, and HKisland scenes) and our newly built \textbf{UAVD4L-2yr}. UAVD4L-2yr pairs an outdated reference map with 8 new query flights ($7.2\text{k}$ frames), introducing two-year seasonal and illumination gaps. It provides centimeter-level RTK-GPS ground truth for the UAV's 6-DoF pose and the corresponding 2D-3D annotations of dynamic targets within the scenes. Further dataset details are provided in \cref{sec:supp_query_data} of the Appendix.

\noindent\textbf{Metrics.}
We evaluate performance using four standard metrics: (1) Median Translation/Rotation Error (m, $^\circ$) for accuracy on successfully localized frames; (2) Recall@{1, 3, 5} (m, $^\circ$), the percentage of frames localized within given translation and rotation thresholds; (3) Completeness (\%), the percentage of frames for which a valid pose is produced without failure; and (4) Frequency (FPS), the rate of localization.

\noindent\textbf{Results.} The results are summarized in \cref{tab:overall_comparison}.
On the synthetic benchmark, PiLoT establishes a new state-of-the-art, outperforming all baselines in localization accuracy and success rate while maintaining the fastest inference speed. \cref{fig:exp_google_error} provides a quantitative, frame-by-frame analysis of localization stability under \textit{Foggy} and \textit{Night} conditions, where PiLoT yields more stable trajectories and lower pose errors. 

On real-world data, deployed without any fine-tuning on the UAVScenes and UAVD4L-2yr, PiLoT again achieves state-of-the-art results with impressive zero-shot generalization. 
For a more in-depth analysis, we provide additional trajectory visualizations under varying motion speeds and weather conditions in the Appendix (see \cref{fig:google_6frames,fig:real_traj,fig:uavscene_6frames}).

\begin{figure*}[t]
\centering
\includegraphics[width=\linewidth]{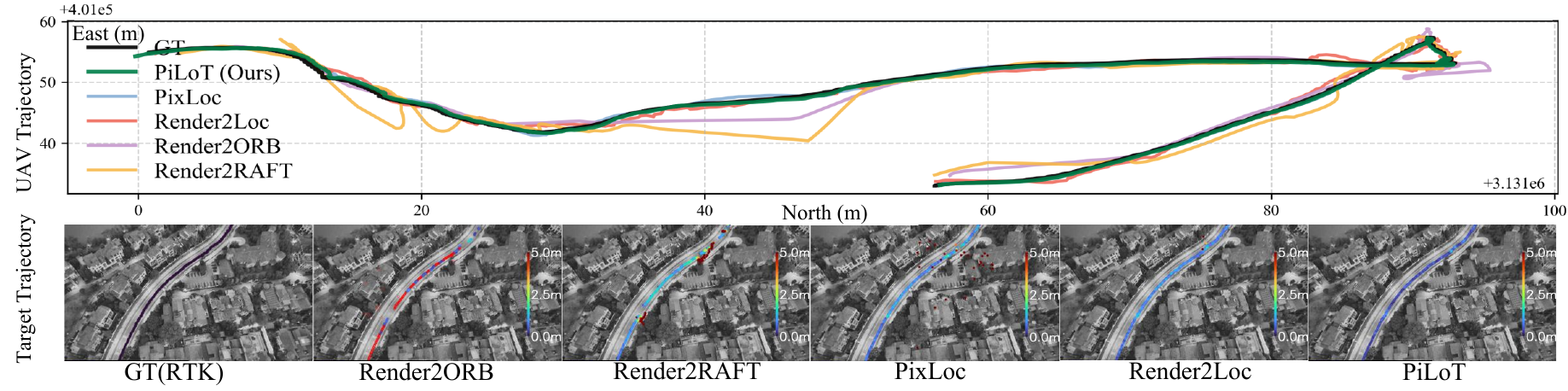} 
\vspace{-6mm}
\caption{
    \textbf{Qualitative results for joint UAV ego localization and dynamic target geo-localization on a UAVD4L-2yr sample sequence.} (Top) Estimated UAV trajectories, where our method (PiLoT, green) most closely follows the ground truth (GT, black). (Middle) The position error of the dynamic target, with color indicating Euclidean distance to the GT. 
}
\label{fig:qualitative_traj}
\vspace{-1mm}
\end{figure*}

\subsection{UAV-based Target Geo-Localization}
\noindent\textbf{Datasets.} We evaluate our method on two benchmarks: a synthetic \textbf{UAVD4L-SynTarget} and a real-world \textbf{UAVD4L-2yr}. UAVD4L-SynTarget is built with UE5 on the UAVD4L scene~\cite{wu2024uavd4l}, including over 100 synthetic targets (vehicles and pedestrians) across 6 UAV trajectories ($6\text{k}$ frames) under diverse weather conditions. It provides precise ground truth for UAV poses, target 2D projections, and 3D geo-locations. Further dataset generation details are provided in \cref{sec:supp_query_data} of the Appendix.
UAVD4L-2yr is the same dataset used for ego-localization, but here we focus on its dynamic target tracking capabilities. Its eight trajectories include dynamic targets with precise annotations, making it ideal for this evaluation.

\noindent\textbf{Metrics.}
We evaluate performance using Recall@$k$ (m), defined as the percentage of targets geolocated within a 3D distance error of $k$ meters from their ground truth position, where $k \in \{1, 3, 5\}$. 

\noindent\textbf{Results.}
We obtain target 3D coordinates via ray casting from the estimated UAV pose, assuming the target is a point-like object with a known 2D pixel location. 
As demonstrated in \cref{tab:real_world_target}, our method's localization translates into state-of-the-art performance, outperforming other methods across both synthetic and real-world scenarios.
This performance gap is visualized qualitatively in \cref{fig:qualitative_traj}. Our method's stable localization thus yields a consistently low-error target trajectory (uniform blue).

\begin{table}[t]
\caption{\textbf{Performance on the dynamic target geo-localization task.} PiLoT's superior ego-localization accuracy translates directly into state-of-the-art target geo-localization performance. We equip Render2loc with the LoFTR matcher, while PixLoc runs with its default 150 iterations.}

\vspace{-2mm}
\scriptsize
\setlength{\tabcolsep}{4pt}
\begin{tabularx}{\linewidth}{@{} l >{\centering\arraybackslash}X >{\centering\arraybackslash}X @{}}
\toprule
& \multicolumn{2}{c}{\textbf{Dynamic Target Indication (R@1/3/5)} $\uparrow$} \\
\cmidrule(lr){2-3}
\textbf{Method} 
& \textbf{Single-Target(Real)} 
& \textbf{Multi-Target(Syn)} \\
\midrule
\multicolumn{3}{@{}l}{\textit{Hybrid Methods (Abs.+Rel.)}} \\
\addlinespace[0.3em]
Render2ORB    & 72.13 / 84.59 / 89.74 & 79.51 / 91.04 / 93.28 \\
Render2RAFT   & 44.15 / 78.96 / 88.19 & 51.33 / 78.12 / 90.50 \\
\addlinespace[0.6em]
\multicolumn{3}{@{}l}{\textit{Absolute Localizers}} \\
\addlinespace[0.3em]
PixLoc        & 83.37 / 87.29 / 91.85 & 86.15 / 91.88 / 93.91 \\
Render2Loc    & 87.62 / 92.60 / 96.25 & 89.03 / 93.15 / 96.07  \\
\rowcolor{ourcolor}
\textbf{PiLoT (Ours)}  & \best{90.81 / 94.32 / 96.85} & \best{93.74 / 95.56 / 98.19}  \\
\bottomrule
\end{tabularx}
\label{tab:real_world_target}
\end{table}

\begin{table}[t]
\centering
\caption{
    \textbf{Ablation study on system components and training data.} 
    We evaluate the contribution of each core component and the effectiveness of our proposed training dataset.
}
\vspace{-3mm}
\scriptsize 
\setlength{\tabcolsep}{4pt} 
\begin{tabularx}{\columnwidth}{
  @{} l   
       *{3}{>{\centering\arraybackslash}X} 
  @{}
}
\toprule
& \multicolumn{3}{c}{\textbf{Recall (\%) @ 1m/1$^{\circ}$ $\uparrow$}} \\
\cmidrule(lr){2-4}
\textbf{Ablation Configuration} & 
  \textbf{\shortstack{w/ 3m, 3$^{\circ}$}} & 
  \textbf{\shortstack{w/ 5m, 5$^{\circ}$}} & 
  \textbf{\shortstack{w/ 10m, 10$^{\circ}$}} \\
\midrule

\multicolumn{4}{@{}l}{\textit{System Components}} \\
\quad Off-the-shelf Backbone & 4.2 & 0.0 & 0.0 \\
\quad + Domain-Specific Training & 51.4 & 43.2 & 15.2 \\
\quad + Rotation-Aware Hypothesis & 83.8 & 78.9 & 70.6 \\
\quad \textbf{+ Motion Regularizer (Ours)} & \textbf{84.3} & \textbf{84.3} & \textbf{84.2} \\
\midrule
\multicolumn{4}{@{}l}{\textit{Training Data Ablation}} \\
\quad Trained on Syn. (no light/weather) & 63.5 & 62.4 & 61.6 \\
\quad Trained on MegaDepth only & 69.9 & 69.5 & 68.7 \\
\quad \textbf{Trained on Ours (Syn. w/ light/weather)} &  \textbf{84.3} & \textbf{84.3} & \textbf{84.2} \\
\bottomrule
\end{tabularx}
\label{tab:ablation_runtime}
\vspace{-5mm}
\end{table}

\subsection{Framework Analysis}
\label{sec:framework_analysis}

\noindent\textbf{Ablation Study.}
\cref{tab:ablation_runtime} summarizes our quantitative findings. Our analysis confirms that domain-specific feature training is foundational. 
The effectiveness of our proposed training data is validated in the \textit{Training Data Ablation}. Training on our synthetic dataset with diverse lighting and weather conditions significantly outperforms both real-world (MegaDepth~\cite{li2018megadepth}) and simpler synthetic baselines. The gain mainly arises from our dataset effectively bridging the domain gap and enabling feature adaptation to the top-down and oblique perspectives typical of UAV imagery.

Building on this foundation, our rotation-aware hypothesis generation and motion regularizer further prove indispensable. Together, they substantially enhance the optimizer’s robustness, enabling stable convergence even under aggressive flight maneuvers.

\noindent\textbf{Runtime Efficiency.} PiLoT achieves real-time performance via a dual-thread architecture and a CUDA-accelerated parallel optimizer. The former provides fresh reference views, while the latter embodies our core strategy of efficiently searching multiple pose hypotheses against a single reference image via massive parallelization. Detailed timing statistics are provided in \cref{sec:dual_thread} of the Appendix.

\noindent\textbf{Limitations and Future Work.} While PiLoT is robust to diverse environmental scenarios, its performance may degrade under extreme visual conditions (e.g., dense fog) or significant calibration bias. Currently, our dependency on high-fidelity 3D models restricts the system's applicability to areas with pre-existing mesh data, hindering wider geographical expansion. To overcome these map acquisition constraints and broaden localization coverage, future work will focus on extending PiLoT to support universal representations, such as Digital Orthophoto Maps (DOM) and Digital Elevation Models (DEM). This transition will facilitate seamless deployment across vast wilderness and urban environments.

\section{Conclusion}

This paper presents PiLoT, a unified framework for UAV ego and target geo-localization through direct registration of video streams to geo-referenced 3D maps. The proposed method introduces three key contributions: 1) a dual-thread architecture that enables real-time performance while maintaining robustness, 2) a large-scale synthetic dataset that facilitates zero-shot sim-to-real feature alignment, and 3) a joint neural-guided stochastic-gradient optimizer that ensures robust convergence under fast motion conditions. Extensive evaluations on both public benchmarks and a newly collected dataset demonstrate that PiLoT achieves state-of-the-art accuracy while operating at over 25 FPS on embedded platforms. We believe this work not only advances the field of vision-based localization for UAVs under GNSS-denied conditions, but also provides valuable insights for localization in other robotic platforms.
\section*{Acknowledgments}
{\small
This research was funded through the Young Scientists Fund of the National Natural Science Foundation of China (NSFC) (Project No. 62406331). 
The authors would like to thank Rouwan Wu, Qing Shuai, Dongli Tan, and Na Zhao for their insightful discussions. 
We sincerely thank \textcolor{pilotpink}{\href{https://cesium.com/platform/cesium-for-unreal/}{Cesium for Unreal}} for providing the data platform and \textcolor{pilotpink}{\href{https://earth.google.com/web/}{Google Earth}} for providing the data source.
}

{
    \small
    \bibliographystyle{ieeenat_fullname}
    \bibliography{main}
}

\clearpage
\appendix
\maketitlesupplementary
\section{Method Details}
\label{Method_details}

\subsection{Network Architecture}
\label{sec:sup_network_arch}
Our feature extraction backbone is a lightweight U-Net architecture, with its detailed data flow outlined in Table~\ref{tab:network_flow}. The encoder path employs the first three stages of a pre-trained MobileOne-S0~\cite{vasu2023mobileone} to efficiently extract a hierarchy of features at strides of 2, 4, and 8. At each stage of the decoder, two parallel heads are applied to generate the multi-level outputs. A projection head, consisting of lightweight 3x3 convolutions, processes the decoder features to produce a feature map. Concurrently, an uncertainty head predicts a per-level single-channel uncertainty map. This entire process yields a three-level pyramid of feature-uncertainty pairs, indexed from coarse to fine ($\ell=0, 1, 2$) corresponding to resolutions of $1/4$, $1/2$, and $1$. 
\begin{table}[h]
\centering
\vspace{-1mm}
\caption{Detailed data flow of our MobileOne-UNet feature extractor. The table illustrates the transformation of tensor shapes through the encoder, decoder, and output heads for an input image of resolution $H \times W = 512 \times 512$.}
\label{tab:network_flow}
\resizebox{\linewidth}{!}{%
\begin{tabular}{@{}llccl@{}}
\toprule
\textbf{Stage} & \textbf{Operation} & \textbf{Resolution} & \textbf{Channels} & \textbf{Description} \\
\midrule
\multicolumn{5}{l}{\textit{\textbf{Encoder (MobileOne-S0)}}} \\
Input & - & $H \times W$ & 3 & Input RGB image \\
Stage 1 ($E_1$) & Stem + Stage 1 & $(H/2) \times (W/2)$ & $C_1$ & First level of features (for skip) \\
Stage 2 ($E_2$) & Stage 2 Blocks & $(H/4) \times (W/4)$ & $C_2$ & Second level of features (for skip) \\
Stage 3 ($E_3$) & Stage 3 Blocks & $(H/8) \times (W/8)$ & $C_3$ & Deepest features fed to decoder \\
\midrule
\multicolumn{5}{l}{\textit{\textbf{Decoder}}} \\
Block 1 ($D_1$) & Upsample($E_3$) + Concat($E_2$) & $(H/4) \times (W/4)$ & 128 & Upsample and fuse with skip connection \\
Block 2 ($D_2$) & Upsample($D_1$) + Concat($E_1$) & $(H/2) \times (W/2)$ & 64 & Upsample and fuse with skip connection \\
Block 3 ($D_3$) & Upsample($D_2$) & $H \times W$ & 32 & Final upsampling (no skip connection) \\
\midrule
\multicolumn{5}{l}{\textit{\textbf{Output Heads (Applied to $D_1, D_2, D_3$)}}} \\
\multirow{2}{*}{Projection} & \multirow{2}{*}{Conv 3x3} & $(H/4) \times (W/4)$ & 32 & \multirow{2}{*}{3-level feature pyramid} \\
 &  & $(H/2) \times (W/2)$ & 32 &  \\
 &  & $H \times W$ & 32 &  \\
\cmidrule{2-5}
\multirow{2}{*}{Uncertainty} & \multirow{2}{*}{Conv 1x1 + Sigmoid} & $(H/4) \times (W/4)$ & 1 & \multirow{2}{*}{3-level confidence maps} \\
 &  & $(H/2) \times (W/2)$ & 1 &  \\
 &  & $H \times W$ & 1 &  \\
\bottomrule
\end{tabular}%
}
\end{table}

\subsection{JNGO Optimizer Details}
\label{sec:sup_lm_formulation}

\begin{figure}[t]
  \centering
  \includegraphics[width=\linewidth]{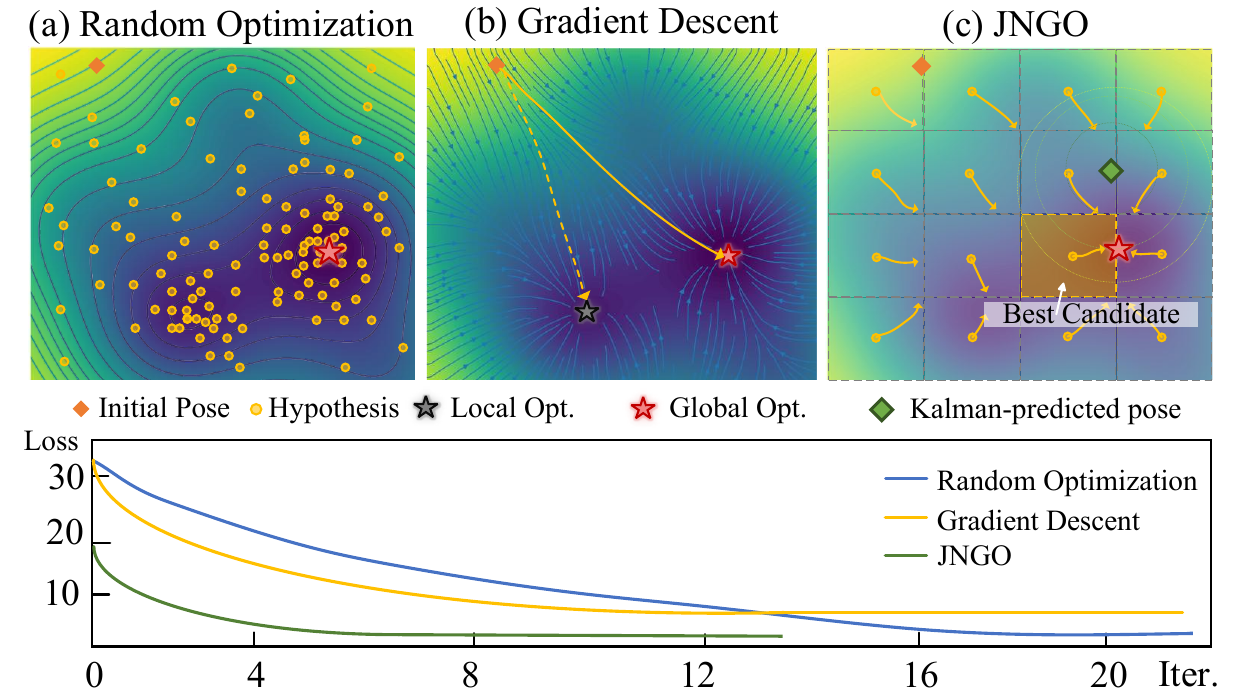}
  \caption{\textbf{Comparison of optimization strategies in a non-convex landscape.} 
While (a) Random Sampling is inefficient and (b) Gradient Descent is prone to local minima, our method (c) efficiently searches for the best solution. The convergence plot below illustrates how our method (green) achieves faster convergence and a lower final loss compared to alternatives.}
\label{fig:optimization_converage}
\end{figure}

As illustrated in ~\cref{fig:optimization_converage}, the feature-metric cost landscape is often highly non-convex, posing a challenge for standard optimization methods. On one hand, exhaustive strategies like random sampling (~\cref{fig:optimization_converage} (a)) are too computationally expensive to be practical. On the other hand, efficient local search methods like gradient descent (~\cref{fig:optimization_converage} (b)) are highly sensitive to initialization and frequently get trapped in suboptimal local minima. To address this, JNGO synergizes stochastic and gradient-based optimization for effective global exploration and local refinement.

\vspace{2mm}
\noindent\textbf{Iterative Linearization and Pose Update.}
The JNGO detailed here serves to minimize the feature-metric cost $\mathcal{C}_{\text{photo}}^{(m, \ell)}$ for each hypothesis, as defined in Eq.6 of the main paper. We solve this non-linear optimization problem iteratively with a small, fixed number of updates per level to maintain real-time performance. Specifically, we perform 2, 3, and 4 LM iterations for the coarse, mid, and fine pyramid levels, respectively. While the full cost includes a robust Huber loss, the LM formulation solves the underlying non-linear least-squares problem based on the residual $\mathbf{r}_{j,\ell}^{(m)}$ from Eq.7.

Each LM iteration solves for a pose increment $\Delta\boldsymbol{\xi} \in \mathfrak{se}(3)$ by linearizing the residual function. For a single residual term $\mathbf{r}_j$ (where $j \in \{1, \dots, N\}$), the linearization at $\tilde{\mathbf{T}}^{(k)}_{m}$ is:
\begin{equation}
\mathbf{r}_j(\exp(\Delta\boldsymbol{\xi}) \cdot \tilde{\mathbf{T}}^{(k)}_{m}) \approx \mathbf{r}_j(\tilde{\mathbf{T}}^{(k)}_{m}) + \mathbf{J}_j \Delta\boldsymbol{\xi},
\end{equation}
where $\mathbf{J}_j$ is the corresponding Jacobian. To solve for the update, we stack the residuals from all $N$ 3D geo-anchors into a single vector $\mathbf{r} = [\mathbf{r}_1^\top, \ldots, \mathbf{r}_\mathbf{N}^\top]^\top$ and their Jacobians into a block matrix $\mathbf{J} = [\mathbf{J}_1^\top, \ldots, \mathbf{J}_\mathbf{N}^\top]^\top$. This yields the normal equations:
\begin{equation}
\left(\mathbf{J}^\top \mathbf{W} \mathbf{J} + \lambda \mathbf{I}\right) \Delta \boldsymbol{\xi} = -\mathbf{J}^\top \mathbf{W} \mathbf{r},
\end{equation}
where $\mathbf{W}$ is a diagonal matrix of learned uncertainty weights. The resulting increment updates the pose $\tilde{\mathbf{T}}^{(k+1)}_{m} = \exp(\Delta \boldsymbol{\xi}) \cdot \tilde{\mathbf{T}}^{(k)}_{m}$.

\begin{figure}[t!]
  \centering
  \includegraphics[width=\linewidth]{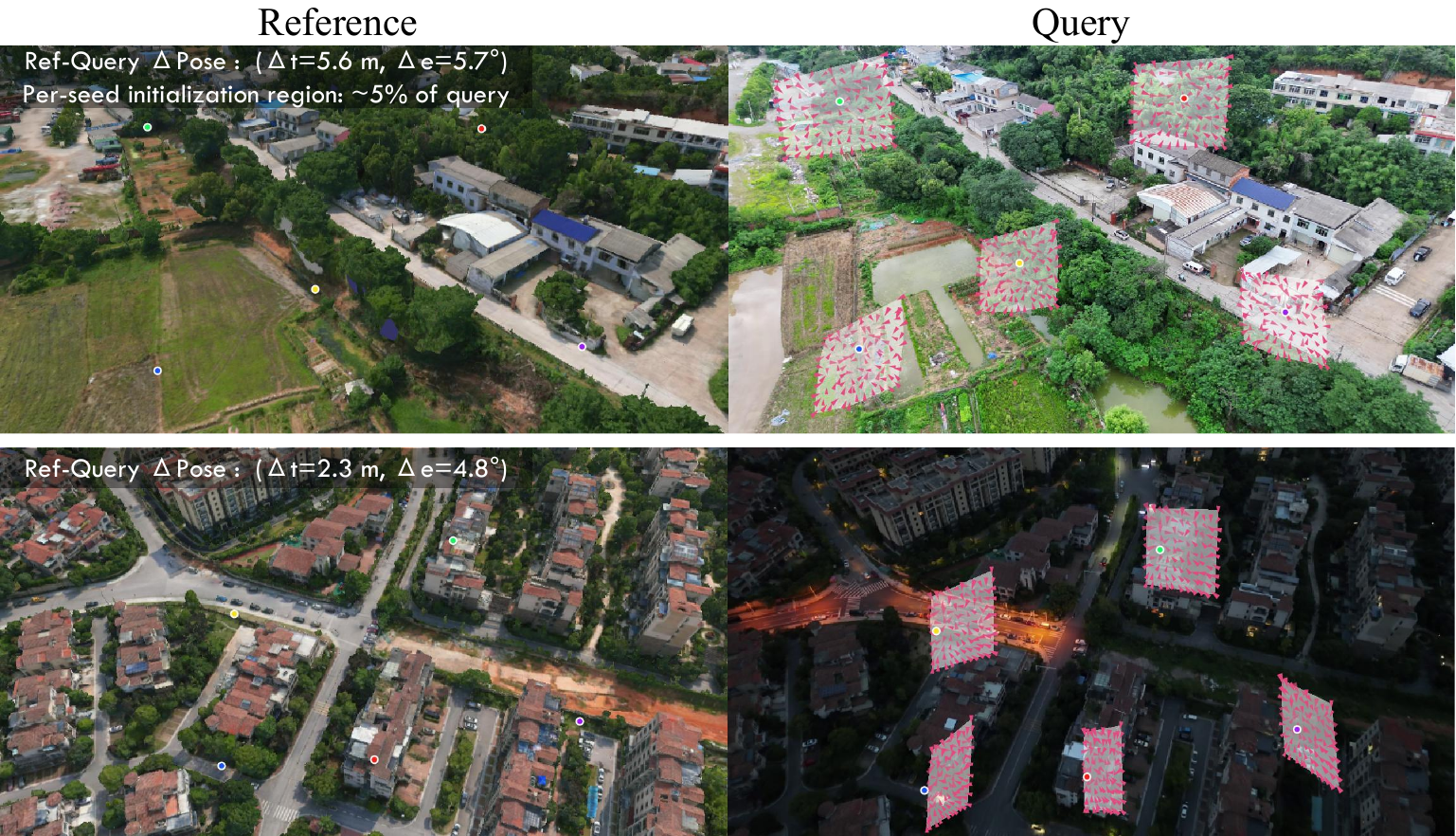}
  \caption{\textbf{Qualitative results of our multi-hypothesis refinement process.} The top row shows a challenging rural scene with significant viewpoint change, while the bottom row demonstrates robustness in a day-night urban setting. The `convergence basin' visualization on the query image (right) demonstrates the subsequent refinement.}
  \label{fig:convergence_basin}
  \vspace{-2mm}
\end{figure}

\begin{figure*}[t]
  \centering
  \includegraphics[width=0.90\linewidth]{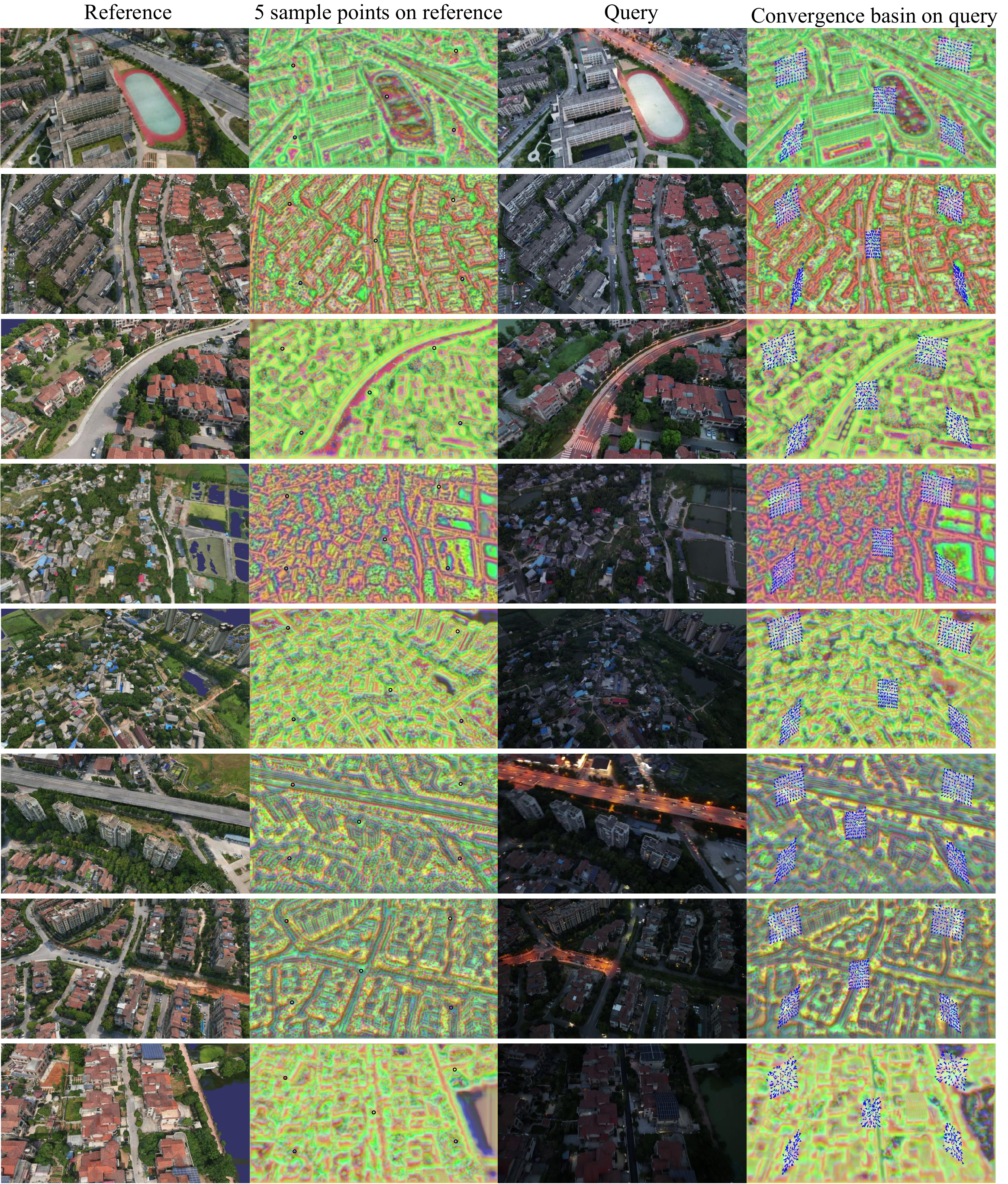}
  \caption{\textbf{Additional examples of the refinement process and its convergence basin.} For various anchor points sampled in the reference image, the initial estimates in the query image (covering a wide area, shown as a field of arrows) are effectively guided by our learned features to converge to a single, precise location.
  }
  \label{fig:feature_coverage}
\end{figure*}

\vspace{2mm}
\noindent\textbf{Jacobian Formulation.}
The Jacobian matrix $\mathbf{J}$ encapsulates the sensitivity of the feature residual to infinitesimal pose perturbations and is derived via the chain rule for each anchor point $j$:
\begin{equation}
\mathbf{J}_j = \frac{\partial \mathbf{r}_j}{\partial \boldsymbol{\xi}} = \underbrace{\frac{\partial \mathbf{f}^q_{\ell}(\tilde{\mathbf{p}}_j^q)}{\partial \tilde{\mathbf{p}}_j^q}}_{\text{Feature Gradient}} \cdot \underbrace{\frac{\partial \pi(\mathbf{P}_j^c)}{\partial \mathbf{P}_j^c}}_{\text{Projection Deriv.}} \cdot \underbrace{\frac{\partial (\tilde{\mathbf{T}}_{m} \mathbf{P}_j^W)}{\partial \boldsymbol{\xi}}}_{\text{Pose Deriv.}}.
\label{eq:sup_jacobian_chain_rule_formal}
\end{equation}
The terms in this chain are:
\begin{itemize}
    \item \textbf{Feature Gradient}: The $(C \times 2)$ matrix $\frac{\partial \mathbf{f}^q_{\ell}}{\partial \tilde{\mathbf{p}}_j^q}$ represents the spatial gradient of the $C$-dimensional query feature map, typically computed using finite differences.
    \item \textbf{Projection Derivative}: The $(2 \times 3)$ matrix $\frac{\partial \pi}{\partial \mathbf{P}_j^c}$ is the standard derivative of the pinhole camera projection with respect to the 3D point coordinates $\mathbf{P}_j^c$ in the camera frame.
    \item \textbf{Pose Derivative}: The $(3 \times 6)$ matrix $\frac{\partial (\tilde{\mathbf{T}}_{m} \mathbf{P}_j^W)}{\partial \boldsymbol{\xi}}$ describes how the 3D point moves in the camera frame as a result of a pose perturbation.
\end{itemize}
The product of these terms yields the final $(C \times 6)$ Jacobian block for a single anchor point, which links the 6-DoF pose update to changes in the feature residual. \Cref{fig:convergence_basin} illustrates that the procedure begins by back-projecting 2D anchor points from the reference image to 3D, and then re-projecting them onto the query image based on initial pose hypotheses. For each seed point (colored dots in the reference image, left), our method initializes a local search region in the query image (right). The red arrows show the model predicting corrective pixel displacements, effectively converging towards the true location. For more convergence examples, please see ~\cref{fig:feature_coverage}.

\begin{algorithm}[ht]
\caption{Fused CUDA Kernel for a Single LM Iteration}
\label{alg:fused_kernel}
\begin{algorithmic}[1]
\State \textbf{Input:} Pose hypotheses $\{\tilde{\mathbf{T}}_m^{(k)}\}$, query features $\mathbf{f}^q$, anchor points $\{\mathbf{P}_j^W\}$, weights $\{w_j\}$
\State \textbf{Output:} System gradient $\mathbf{g}$ and Hessian $\mathbf{H}$ for each hypothesis
\ForAll{hypothesis $\tilde{\mathbf{T}}_m^{(k)}$ in parallel}
    \State Initialize global gradient $\mathbf{g} \in \mathbb{R}^6$ and Hessian $\mathbf{H} \in \mathbb{R}^{6\times6}$ to zero.
    \State \textbf{Launch one CUDA thread per anchor point $\mathbf{P}_j^W$}
    \State \texttt{// --- In-thread computation (on-chip registers) ---}
    \State Project point: $\mathbf{P}_j^c = \tilde{\mathbf{T}}_m^{(k)} \mathbf{P}_j^W$, then $\tilde{\mathbf{p}}_j^q = \pi(\mathbf{K}, \mathbf{P}_j^c)$
    \State Compute residual $\mathbf{r}_j = \mathbf{f}^q(\tilde{\mathbf{p}}_j^q) - \mathbf{f}^r(\mathbf{p}_j^r)$
    \State Compute Jacobian $\mathbf{J}_j$ via chain rule (~\cref{eq:sup_jacobian_chain_rule_formal})
    \State Compute per-point contribution: $\mathbf{g}_j = \mathbf{J}_j^\top w_j \mathbf{r}_j$ and $\mathbf{H}_j = \mathbf{J}_j^\top w_j \mathbf{J}_j$
    \State \texttt{// --- Global memory reduction ---}
    \State Update global system using \texttt{atomicAdd}:
    \State $\mathbf{g} \gets \mathbf{g} + \mathbf{g}_j$
    \State $\mathbf{H} \gets \mathbf{H} + \mathbf{H}_j$
    \State \textbf{end parallel for} (threads sync implicitly)
\EndFor
\State Solve $(\mathbf{H} + \lambda\mathbf{I})\Delta\boldsymbol{\xi} = -\mathbf{g}$ for all hypotheses on GPU.
\end{algorithmic}
\end{algorithm}

\vspace{1mm}
\noindent\textbf{Rotation-Aware Sampling Strategy.} 
To justify our sampling design, we analyze the 6-DoF convergence basin of the proposed optimizer against the physical constraints of high-speed UAV motion. 
As reported in Table~\ref{tab:conv_limits}, we contrast our convergence limits with the maximum inter-frame motion (at 30\,fps) derived from DJI Matrice 4 specifications. 
Our analysis reveals that while translation ($T_x, T_y, T_z$) and gimbal-stabilized Roll ($\phi$) axes stay well within the convergence bounds (marked in \textcolor{green!60!black}{green}), the out-of-plane Yaw ($\psi$) and Pitch ($\theta$) motions frequently exceed the initial convergence limits ($3.5^\circ < 6.7^\circ$ and $2.7^\circ < 3.0^\circ$, respectively). 
To bridge this gap and ensure robustness during aggressive maneuvers, we specifically expand the search space for Yaw and Pitch via the proposed \textit{Rotation-Aware Sampling} strategy, effectively broadening the basin for these critical axes.

\begin{figure}[t!]
    \centering
    \includegraphics[width=\linewidth]{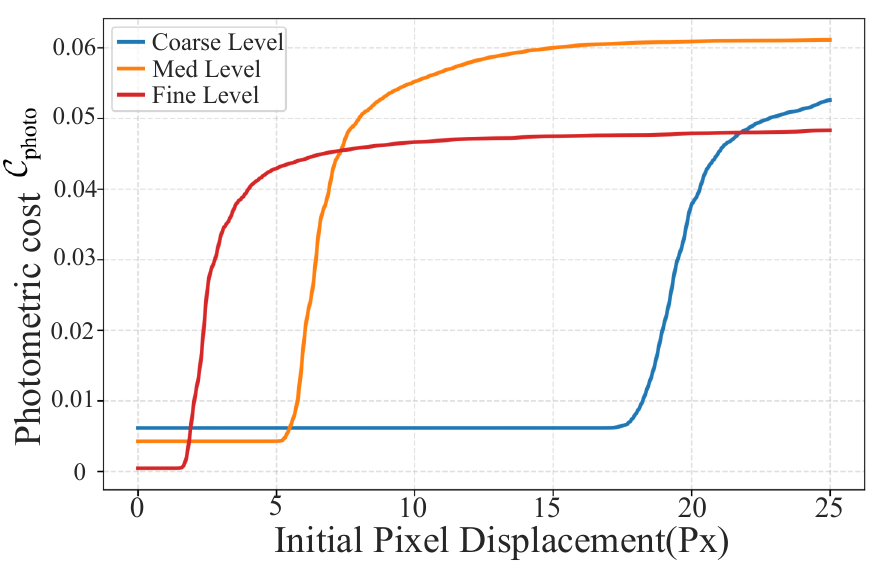}
    \caption{Convergence basin evolution across levels.}
    \label{fig:c2f_analysis}
\end{figure}

\begin{table}[h]
    \centering
    \caption{Comparison of 6-DoF convergence limits vs. maximum inter-frame motion (based on DJI specs at 30\,fps).}
    \label{tab:conv_limits}
    \scriptsize
    \setlength{\tabcolsep}{4pt}
    \begin{tabular}{l | c c c | c c c}
    \toprule
    Metric & \textbf{Yaw} ($\psi$) & \textbf{Pitch} ($\theta$) & \textbf{Roll} ($\phi$) & \textbf{Tx} & \textbf{Ty} & \textbf{Tz} \\
    \midrule
    Conv. Limit & 3.5$^{\circ}$ & 2.7$^{\circ}$ & 3.2$^{\circ}$ & 5.6m & 8.1m & 7.4m \\
    Max Motion (30 fps) & \textcolor{red}{6.7$^{\circ}$} & \textcolor{red}{3.0$^{\circ}$} & \textcolor{green!60!black}{<0.1$^{\circ}$} & \textcolor{green!60!black}{0.7m} & \textcolor{green!60!black}{0.7m} & \textcolor{green!60!black}{0.3m} \\
    \bottomrule
    \end{tabular}
\end{table}
\noindent\textbf{Coarse-to-Fine Analysis.} 
We further investigate the evolution of the convergence basin across different optimization scales. 
As illustrated in Fig.~\ref{fig:c2f_analysis}, the convergence basin exhibits a characteristic narrowing as the precision increases (Coarse $\rightarrow$ Mid $\rightarrow$ Fine). 
While the \textit{Fine} level provides high-precision localization, its limited basin is often insufficient to capture large inter-frame displacements. 
Conversely, the \textit{Coarse} level offers a significantly broader basin, providing the necessary robustness to large initial pose errors. 
This hierarchical behavior validates our coarse-to-fine design: the coarse stage provides a robust initialization that brings the pose within the tight convergence range of the fine stage, ultimately achieving a balance between global robustness and local precision.

\noindent\textbf{High-Performance CUDA Implementation with Kernel Fusion.}
\label{sec:sup_cuda_impl}
To achieve real-time performance, the computationally intensive JNGO is accelerated with a custom CUDA implementation. A naive GPU port would involve multiple kernel launches for each step of the Jacobian calculation (e.g., projection, feature gradient sampling, matrix products), leading to significant memory bandwidth bottlenecks and launch overhead. Our key optimization is the use of a single, highly-fused CUDA kernel. As outlined in ~\cref{alg:fused_kernel}, we parallelize the computation by launching one GPU thread for each of the $J$ anchor points. Each thread is responsible for the entire calculation chain for its assigned point:
\begin{enumerate}
    \item Projecting the 3D world point $\mathbf{P}_j^W$ into the query camera frame.
    \item Calculating the full Jacobian chain $\mathbf{J}_j = \frac{\partial \mathbf{f}^q}{\partial \mathbf{p}^q} \frac{\partial \pi}{\partial \mathbf{P}^c} \frac{\partial (\mathbf{T}\mathbf{P}^W)}{\partial \boldsymbol{\xi}}$ and the feature residual $\mathbf{r}_j$. All intermediate values are kept within fast on-chip registers.
    \item Computing the per-point contribution to the final system: the gradient vector $\mathbf{g}_j = \mathbf{J}_j^\top \mathbf{W}_j \mathbf{r}_j$ and the Hessian matrix $\mathbf{H}_j = \mathbf{J}_j^\top \mathbf{W}_j \mathbf{J}_j$.
    \item Atomically adding these local contributions to the global gradient vector $\mathbf{g}$ and Hessian matrix $\mathbf{H}$ in global memory.
\end{enumerate}

This strategy minimizes costly global memory access, maximizing computational throughput. Once the kernel completes, the final small ($6 \times 6$) system of normal equations is solved in a batch for all $M$ hypotheses using a parallel Cholesky decomposition on the GPU. As shown in ~\cref{tab:cuda_performance}, our fused CUDA kernel dramatically reduces latency, achieving over 30x speedup, making our wide-area parallel search feasible in real-time.

\begin{table}[t]
\centering
\caption{\textbf{Latency Comparison for a Single LM Iteration.} We compare different implementations for computing the cost, gradient, and Hessian. Our fused CUDA kernel dramatically reduces latency, achieving over 30x speedup compared to the initial ONNX-based implementation.}
\label{tab:cuda_performance}
\resizebox{\columnwidth}{!}{%
\begin{tabular}{lccccr}
\toprule
\textbf{Image Size} & \textbf{ONNX Runtime} & \textbf{C++ Reference} & \textbf{CUDA Port} & \multicolumn{2}{c}{\textbf{CUDA Fused Kernel } \adjustbox{valign=c}{\includegraphics[height=1.2\ht\strutbox]{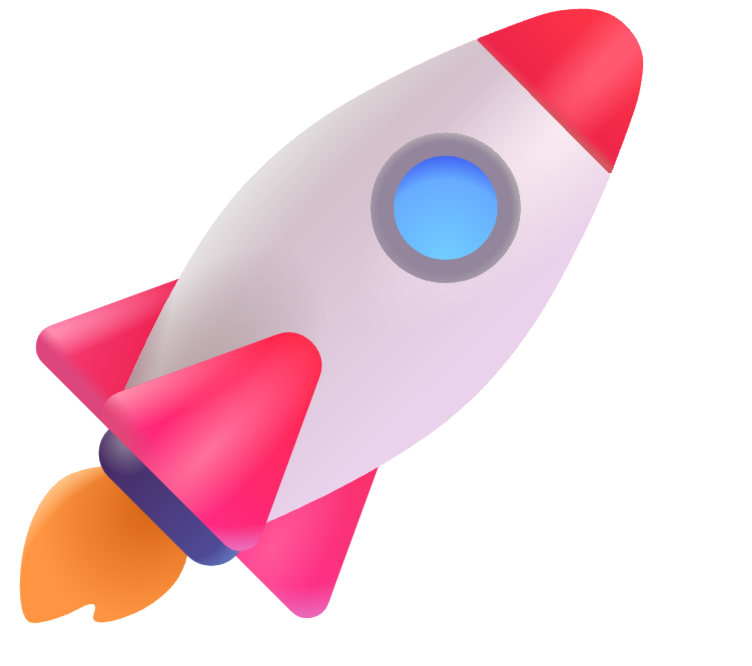}}} \\
\midrule
512 $\times$ 512 & 14.9 ms & 8.44 ms & 5.72 ms & \multicolumn{2}{c}{\textbf{0.49 ms} \adjustbox{valign=c}{\includegraphics[height=1.2\ht\strutbox]{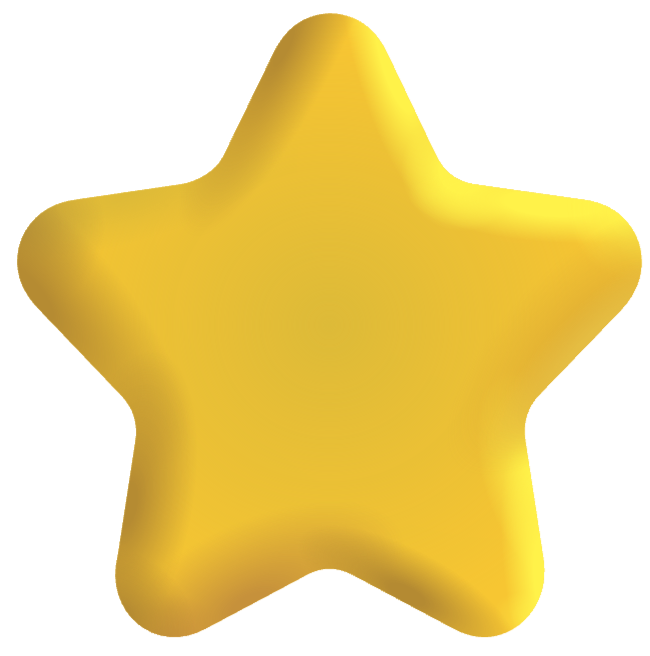}}} \\
256 $\times$ 256 & 10.4 ms & 8.35 ms & 5.24 ms & \multicolumn{2}{c}{\textbf{0.47 ms}  \adjustbox{valign=c}{\includegraphics[height=1.2\ht\strutbox]{figs/star.png}}} \\
128 $\times$ 128 & 8.8 ms  & 8.27 ms & 5.21 ms & \multicolumn{2}{c}{\textbf{0.44 ms}  \adjustbox{valign=c}{\includegraphics[height=1.2\ht\strutbox]{figs/star.png}}} \\
\bottomrule
\end{tabular}%
}
\end{table}

\subsection{Dual-Thread Synchronization}
\label{sec:dual_thread}

\begin{figure}[h]
    \centering
    \includegraphics[width=\columnwidth]{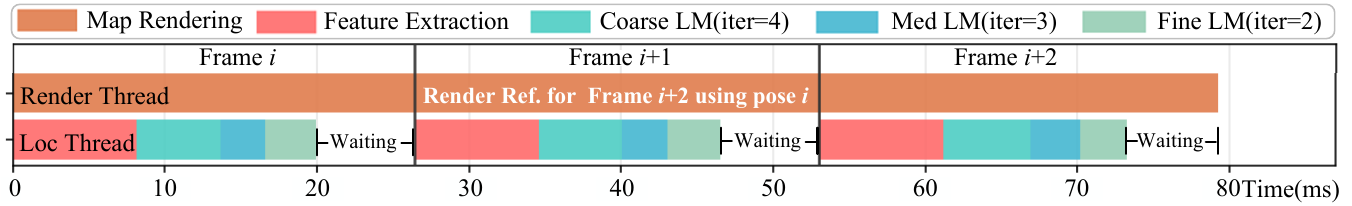}
    \caption{\textbf{Timing diagram of the dual-thread synchronization.} The pipeline maintains a fixed 2-frame lag, which is compensated by motion extrapolation to ensure real-time accuracy.}
    \label{fig:rebuttal_timing}
    \vspace{-4mm}
\end{figure}

Beyond low-level kernel optimization, PiLoT employs a synchronized dual-thread architecture to manage the interplay between map rendering and pose estimation. As detailed in Fig.~\ref{fig:rebuttal_timing}, our pipeline ensures strict temporal alignment: a new localization cycle only commences after the preceding rendering task is completed. This design maintains a consistent 2-frame lag between the reference view and the current query frame, which is essential for stable GPU memory management during high-speed rendering.

To mitigate the resulting latency (approx. 25\,ms), we incorporate a constant-velocity motion model that extrapolates the UAV's pose to the exact rendering timestamp. Given our system operates at video-rate ($<40$\,ms intervals), the pose drift during this brief lag remains minimal. 

\section{Dataset Details}
\label{sec:supp_dataset_details}

\subsection{Large-Scale Synthetic Dataset}
\label{sec:supp_data_generation}

In this section, we provide a detailed overview of our synthetic dataset, including the generation pipeline, quality validation procedures, and comprehensive statistics.

\begin{figure}[t!]
  \centering
  \includegraphics[width=\linewidth]{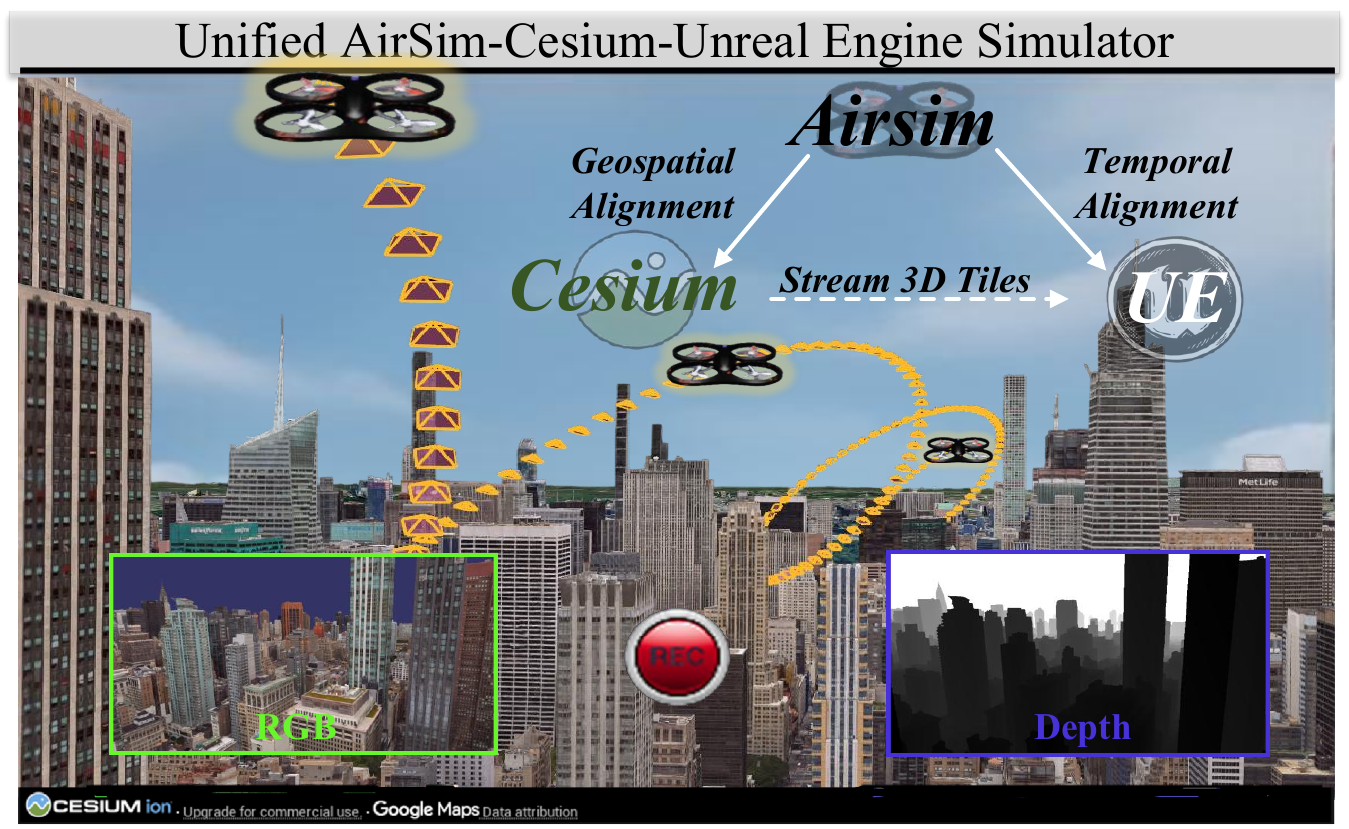}
  \caption{\textbf{The AirSim-Cesium-Unreal Engine Simulator Interface.} Our system integrates Unreal Engine (UE) for real-time rendering, Cesium to load Google 3D Tiles models, and AirSim to simulate UAV flight missions. This pipeline enables the synchronous acquisition of photo-realistic query images and accurate 6-DoF ground truth poses within large-scale, geo-referenced environments.}
\label{fig:system_overview_spacious}
\vspace{-2mm}
\end{figure}

\noindent\textbf{Data Generation Workflow.}
We developed a fully automated data acquisition pipeline, as shown in ~\cref{fig:system_overview_spacious}. This pipeline is engineered specifically to overcome the common geospatial and temporal incoherence in simulation-based data generation, ensuring high spatio-temporal coherence. The core components of our system include:
\begin{enumerate}
\item Geospatial Foundation: Leveraging \textbf{Cesium for Unreal}, which streams high-resolution photogrammetric 3D Tiles from Google, to provide a high-fidelity digital twin of the real world.
\item Flight and Sensor Dynamics: Utilizing \textbf{AirSim} for simulating a multi-rotor UAV, enabling precise and reproducible trajectory execution based on WGS84 waypoints.
\item Photorealistic Rendering: Harnessing \textbf{Unreal Engine}'s physically-based rendering engine to simulate a wide array of lighting and weather conditions.
\end{enumerate}

\begin{figure}[b]
  \centering
  \includegraphics[width=\linewidth]{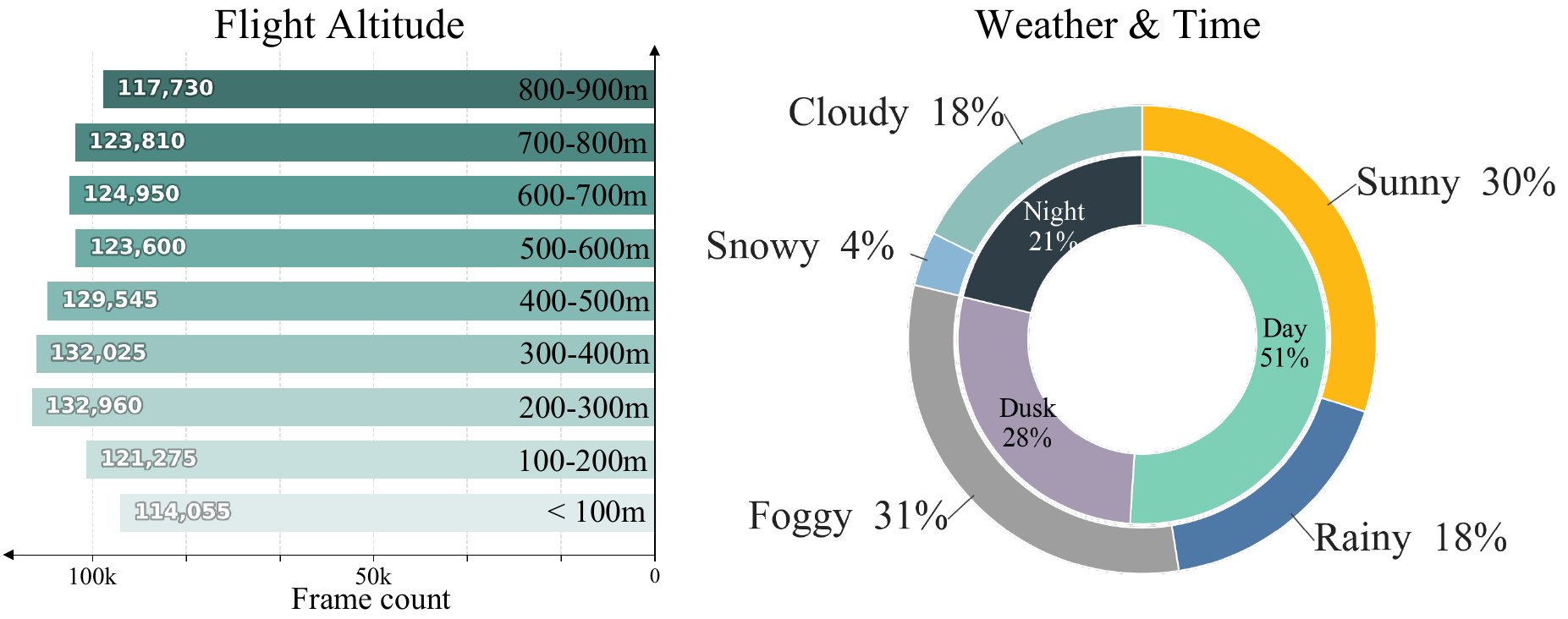}
  \caption{Dataset statistics for flight altitude (left) and environmental conditions with time of day (right).}
  \label{fig:data_statistics}
  \vspace{-2mm}
\end{figure}

\begin{figure}[t]
  \centering
  \includegraphics[width=\linewidth]{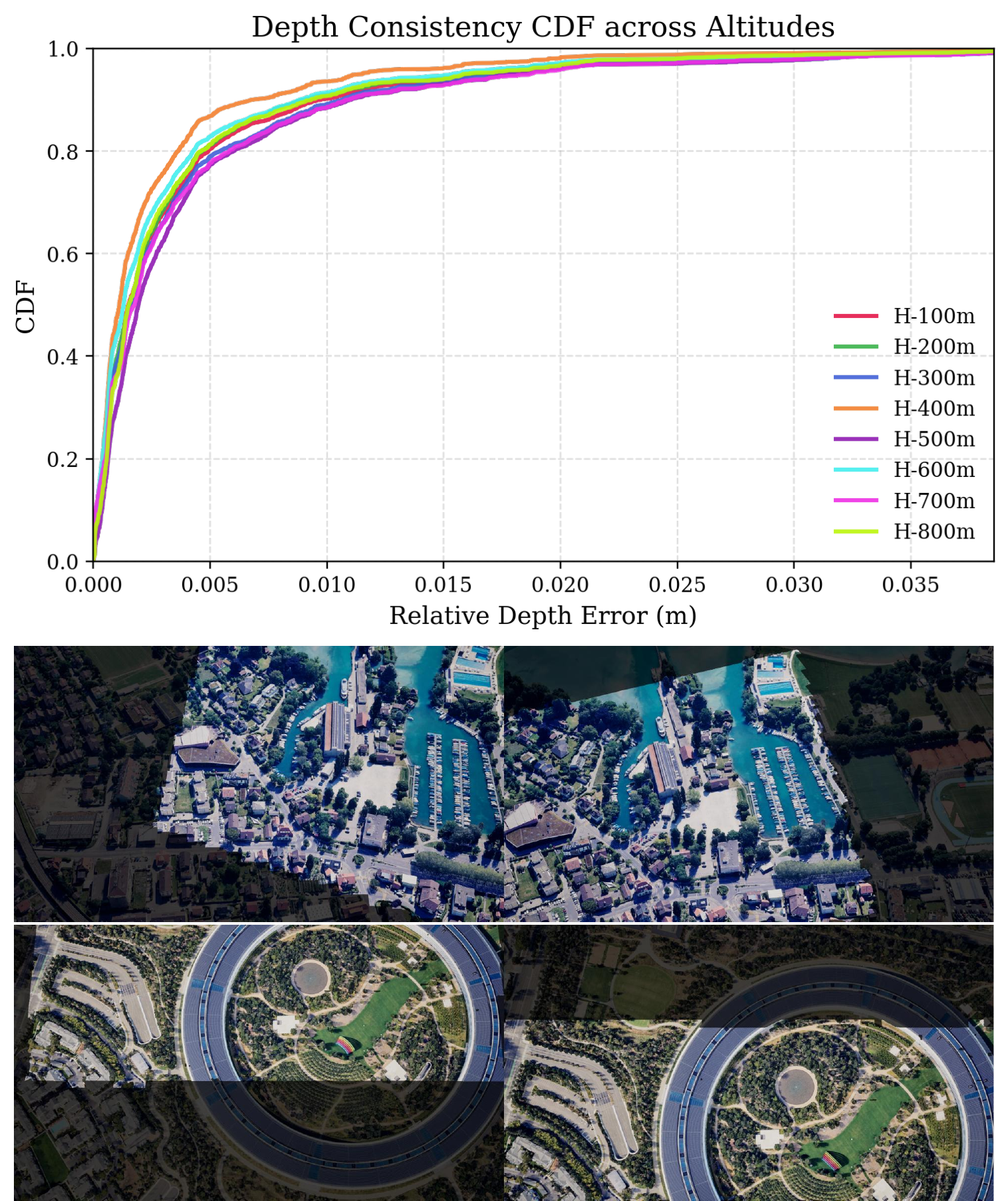}
  \caption{\textbf{Geometric Consistency Validation of Synthetic Data.} \textbf{Top:} The Cumulative Distribution Function (CDF) of bidirectional relative depth errors across different altitudes (100m--800m). \textbf{Bottom:} Qualitative alignment results. The middle and bottom row illustrates the accurate mapping of co-visible regions across disparate perspectives.}
  \label{fig:supp_geometric_validation}
\end{figure}

\vspace{2mm}
\noindent\textbf{Dataset Statistics and Diversity.}
Our dataset comprises over 1.1 million rendered images from 82 diverse regions, captured across over 650 km of UAV flight trajectories and featuring a rich mix of urban and natural landscapes. Environmental conditions are systematically varied across multiple weather conditions (Sunny, Cloudy, Rainy, Foggy, Snowy) and times of day (Day, Sunset, Night), and each region includes four trajectories rendered under distinct, randomly sampled weather settings to support cross-condition training and evaluation. The detailed distribution of these conditions and flight altitudes is summarized in ~\cref{fig:data_statistics}.
Our acquisition flights cover the sub-800 m UAV operating envelope, and the camera pitch is swept from 20° (oblique) to 90° (nadir) to approximate arbitrary viewpoints and enable rigorous evaluation of robustness to viewpoint changes. 

All images are rendered at a resolution of 1600$\times$1200. We use a pinhole camera model with per-frame intrinsics (\textit{$f_x, f_y, c_x, c_y$}). For each frame, we provide the full 6-DoF pose, where the position is given in both WGS84 (longitude, latitude, height) and ECEF formats (X, Y, Z), and the rotation is provided as Euler angles (pitch, roll, yaw). This representation allows for seamless conversion to local project frames (e.g., for COLMAP or OSG) through well-defined transforms.

\begin{figure}[t!]
  \centering
  \includegraphics[width=\linewidth]{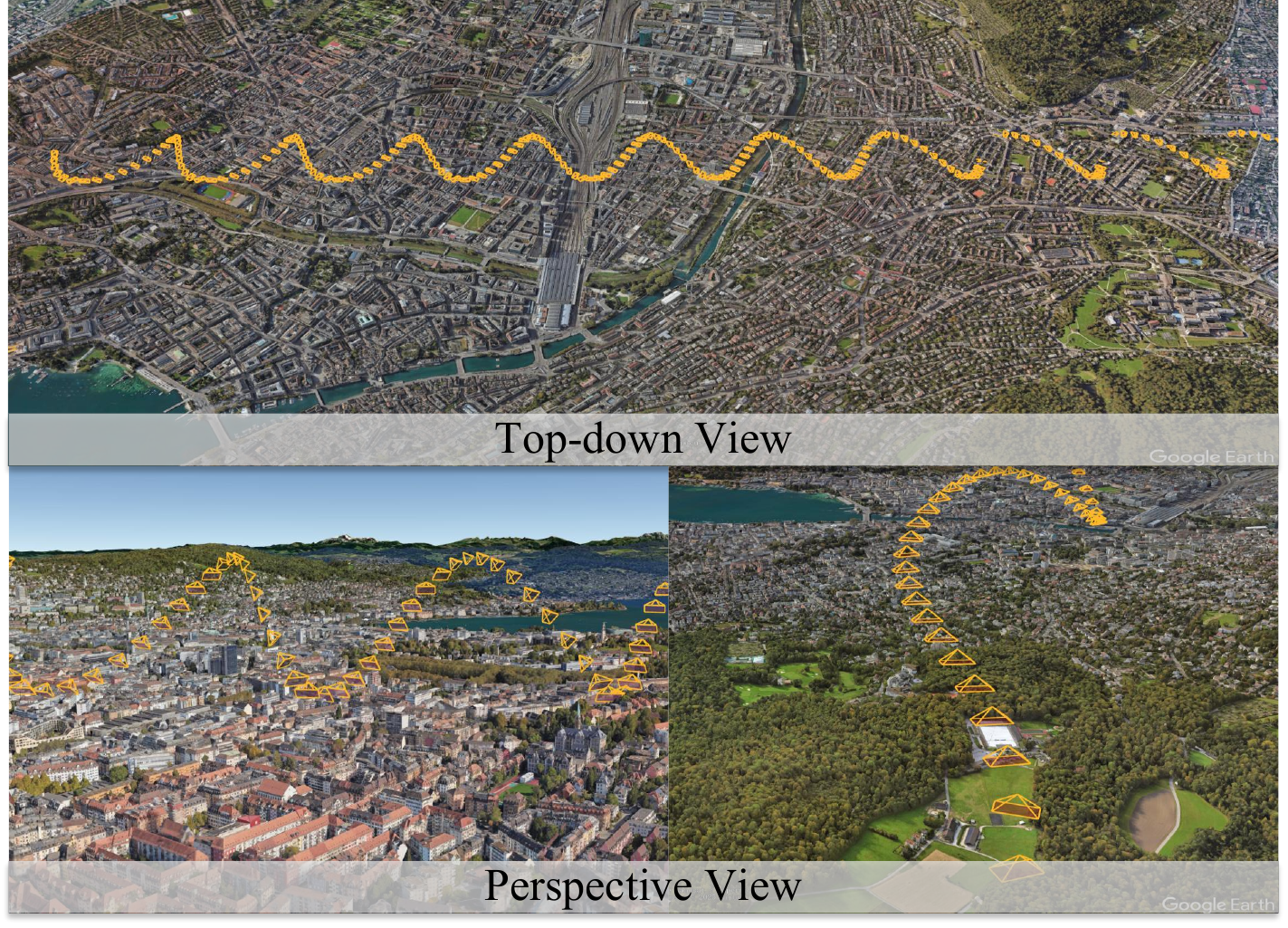}
  \caption{\textbf{Visualization of the designed "barrel-roll" trajectory.} The top-down view (top) shows yaw variation along the S-shaped path. The perspective views (bottom) illustrate changes in altitude (up-and-down arcs) and pitch (tilting of camera frustums). }
  
  \label{fig:supp_flight_design}
\end{figure}
\vspace{2mm}
\noindent\textbf{Data Quality and Validation.}
To ensure the geometric fidelity of our synthetic benchmark, we conduct both quantitative and qualitative validations. 
As illustrated in ~\cref{fig:supp_geometric_validation}, the Cumulative Distribution Function (CDF) of bidirectional relative depth errors across diverse altitudes (100m--800m) demonstrates high numerical precision, with over 90\% of pixels exhibiting a relative depth error of less than 0.01 m. 
Qualitatively, depth-based warping results show seamless pixel-level alignment and consistent projection of co-visible regions across disparate viewpoints. 

\vspace{2mm}
\noindent\textbf{Flight Trajectories Generation.} 
We generate a set of “barrel-roll–inspired” style flight trajectories for data collection, as shown in ~\cref{fig:supp_flight_design}. These trajectories combine horizontal orbiting around a central axis at a fixed radius with step-wise adjustments in \textit{pitch}, \textit{yaw} angle and \textit{altitude}. The flight paths also integrate straight-line cruising and curved sweeps to introduce sufficient translational and viewpoint changes, providing the necessary data for training models to be robust against cross-view and cross-scale challenges.
\begin{figure}[t!]
  \centering
  \includegraphics[width=\linewidth]{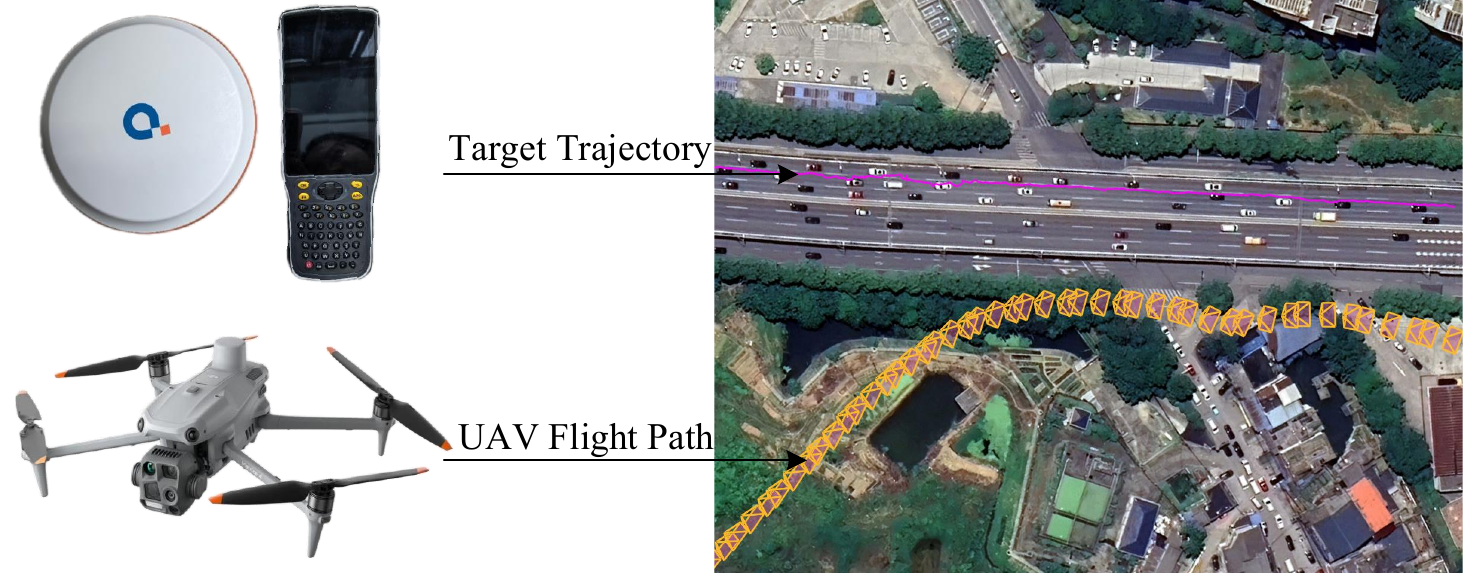}
  \caption{\textbf{Data collection setup and trajectory visualization.} 
(Left) The hardware used for data acquisition: a DJI M4T drone for capturing query images and Qianxun RTK instruments for ground truth positioning. 
(Right) A top-down view of the UAV flight path (query trajectory, shown in orange) and the corresponding ground truth (target trajectory, shown in purple).}

  \label{fig:supp_dataset_trajectories}
\end{figure}

\subsection{Evaluation Datasets}
\label{sec:supp_query_data}
\begin{figure*}[t]
  \centering
  \includegraphics[width=0.90\textwidth]{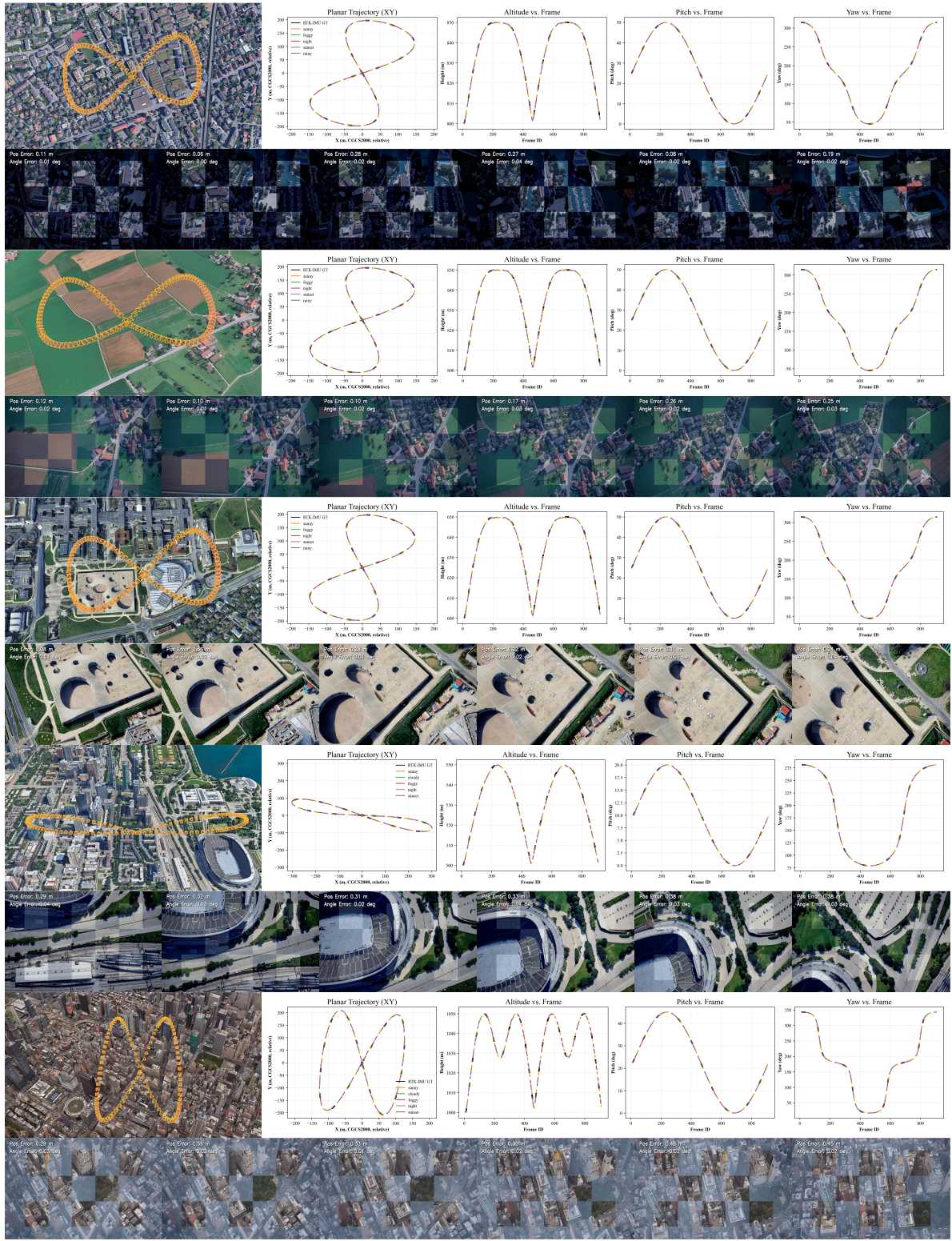}
  \caption{\textbf{Trajectory estimation results on various synthetic scenes from the \textit{SynthCity-6} dataset.} The figure showcases our method's performance across diverse synthetic conditions, including night (\textit{Switzerland-seq4, -seq7}), cloudy (\textit{Switzerland-seq12}), sunny (\textit{USA-seq2}), and foggy (\textit{USA-seq5}).}
  \label{fig:google_6frames}
\end{figure*}

\begin{figure*}[t]
  \centering
  \includegraphics[width=0.90\linewidth]{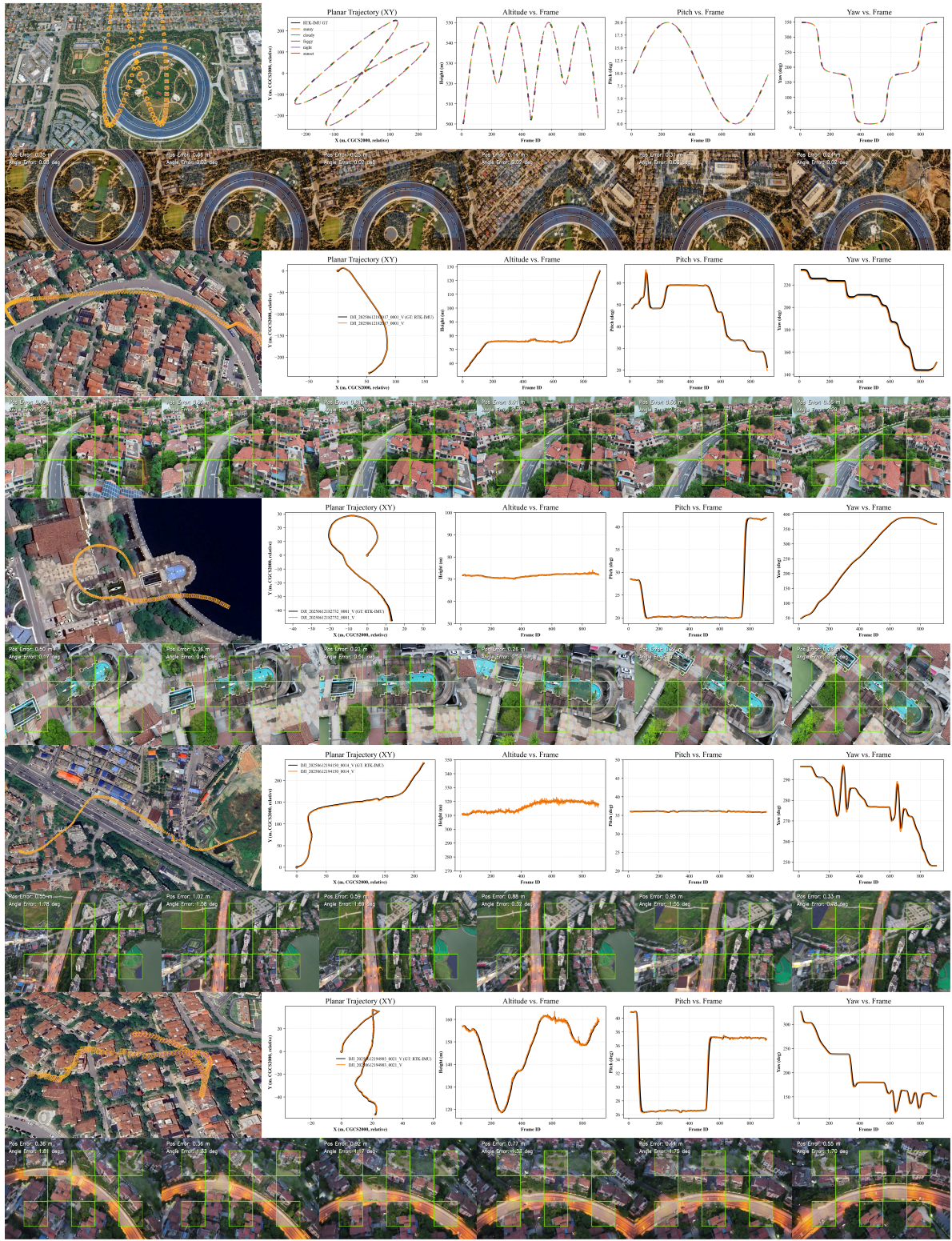}
   \caption{\textbf{Trajectory estimation results on challenging real-world and long-term scenarios.} This figure demonstrates the model's robustness by evaluating on: (1) a synthetic sunset scene from \textit{SynthCity-6 (USA-seq8)}, and (2) 4 real-world sequences from the \textit{UAVD4L-2yr} dataset, which include challenging day/night conditions (\textit{seq2, seq3, seq6, seq8}).}
  \label{fig:real_traj}
\end{figure*}

\noindent\textbf{SynthCity-6 Dataset.}
We construct a synthetic test set, SynthCity-6, using the same data generation workflow as our training set. The test set uses six new locations from different regions of Switzerland and the USA, ensuring no geographic overlap with the training data. As detailed in ~\cref{tab:SynthCity-6_summary}, each sequence is rendered under 5 weather/illumination conditions (sunny, sunset, night, foggy, cloudy/rainy) and at two different altitudes (200m and 500m) to introduce significant scale variations. In total, SynthCity-6 contains 54,000 camera frames with synchronized 6-DoF poses, creating a comprehensive and challenging benchmark for assessing model robustness. For detailed trajectory visualizations and localization results on this dataset, please refer to~\cref{fig:google_6frames} and ~\cref{fig:real_traj}.

\begin{figure}[t!]
  \centering
  \includegraphics[width=\linewidth]{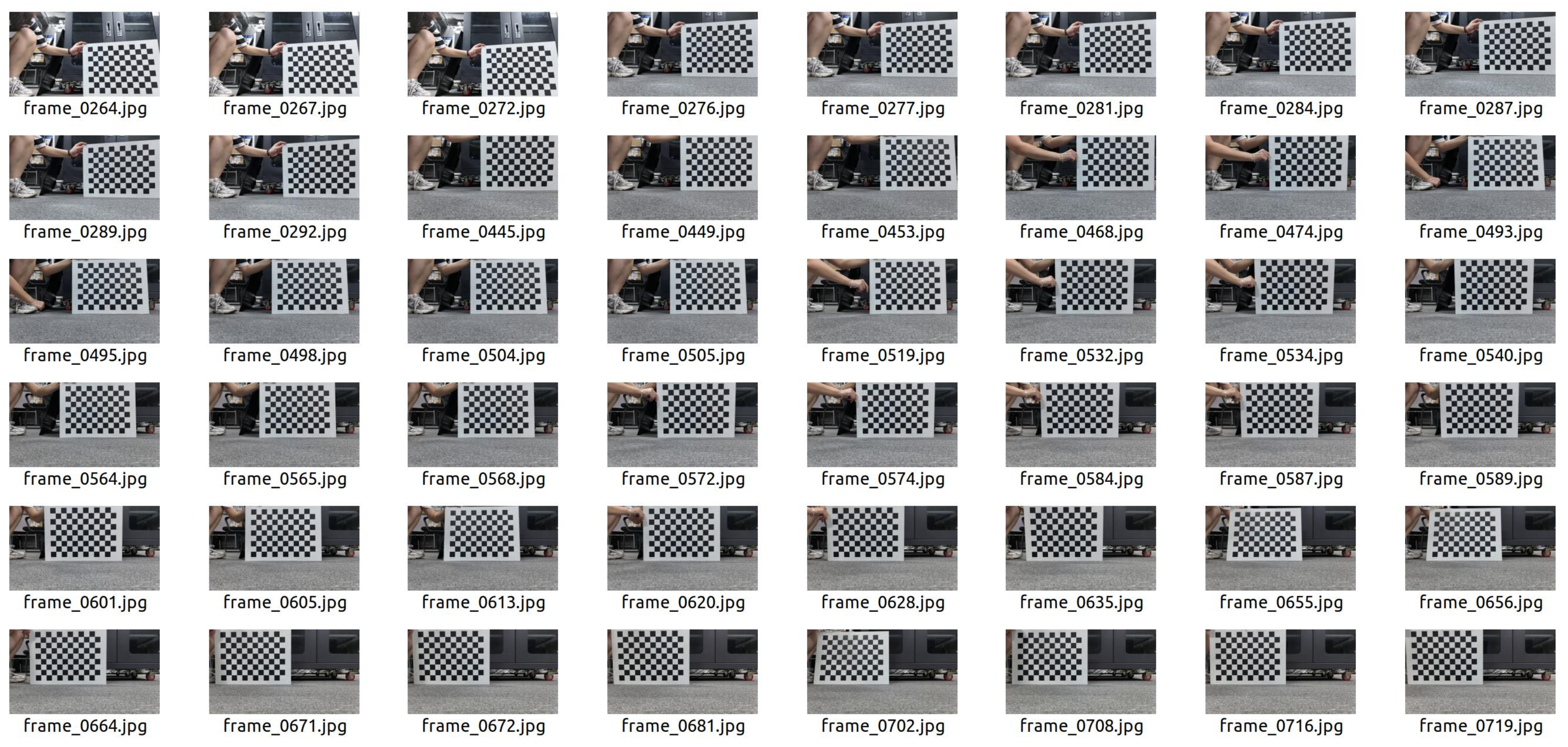}
  \caption{Sample checkerboard images used for intrinsic calibration.}
  \label{fig:supp_camera_intrinsic}
\end{figure}

\vspace{2mm}
\noindent\textbf{UAVD4L-2yr Dataset.}
We build a new dataset for UAV-based ego and target geo-localization. For the reference map, we utilize the publicly available UAVD4L dataset~\cite{wu2024uavd4l}, which covers a large-scale urban area (~$100,000~\text{m}^2$) containing diverse buildings, streets, and vegetation. Our query sequences are captured using a DJI Matrice 4T (M4T) drone~\footnote{https://enterprise.dji.com/cn/matrice-4-series}. As detailed in ~\cref{tab:UAVD4L-2yr_summary}, we collect data under varying illuminations (day and night), different environments (dense urban and sparse rural areas), and dynamic scenes with moving objects. Notably, a two-year time gap separates the query data from the reference map, posing a significant challenge due to long-term appearance variations. The flight paths and the target's trajectory are visualized in ~\cref{fig:supp_dataset_trajectories}. 

The drone is equipped with a centimeter-level RTK module and a high-precision IMU to record its 6-DoF ground-truth pose. Concurrently, the ground-truth poses of the moving target were independently captured using a handheld Qianxun RTK device~\footnote{https://en.qxwz.com/en-product/Q300}. Both the drone's and the target's RTK systems were synchronized to a common UTC time source. All ground-truth poses were subsequently transformed into a unified ECEF coordinate system to ensure precise spatio-temporal alignment between the query and target trajectories. The drone's video camera intrinsics were pre-calibrated using Zhang's method, as detailed in ~\cref{fig:supp_camera_intrinsic}. For detailed trajectory visualizations and localization results on this dataset, please refer to~\cref{fig:real_traj}.

\begin{figure*}[t]
  \centering
  \includegraphics[width=0.90\textwidth]{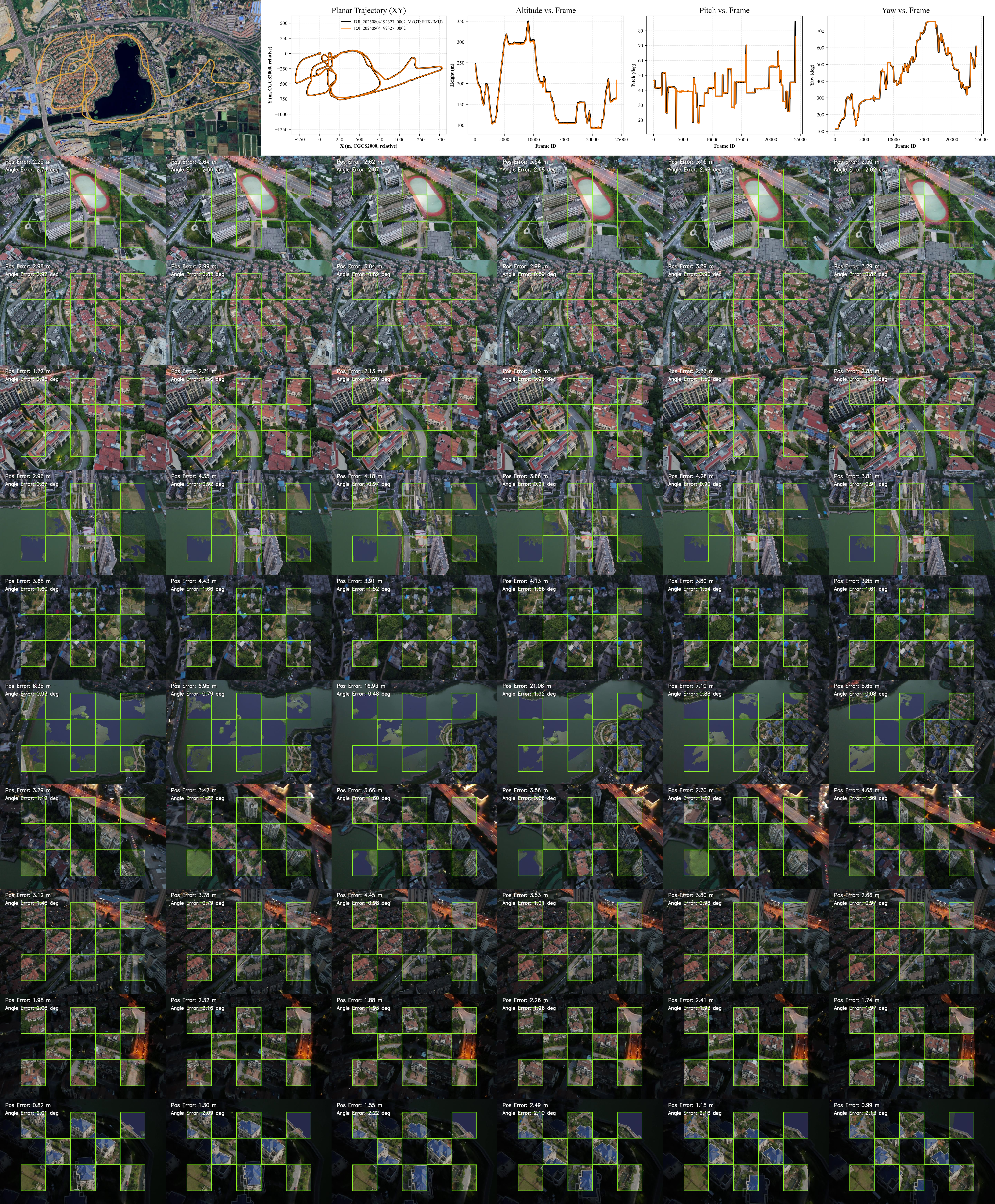}
  \caption{\textbf{Trajectory estimation results on \textit{Long Trajectory Flights}.} We compare our method's estimated trajectory (orange) against the ground truth from an RTK-IMU system (black). The plots show high consistency for planar position (XY), altitude, pitch, and yaw. The checkerboard insets provide an AR visualization, which overlays the live camera views with the rendered views based on our estimated pose.}
  \label{fig:long_6frames}
\end{figure*}

\begin{figure*}[t]
  \centering
  \includegraphics[width=0.90\linewidth]{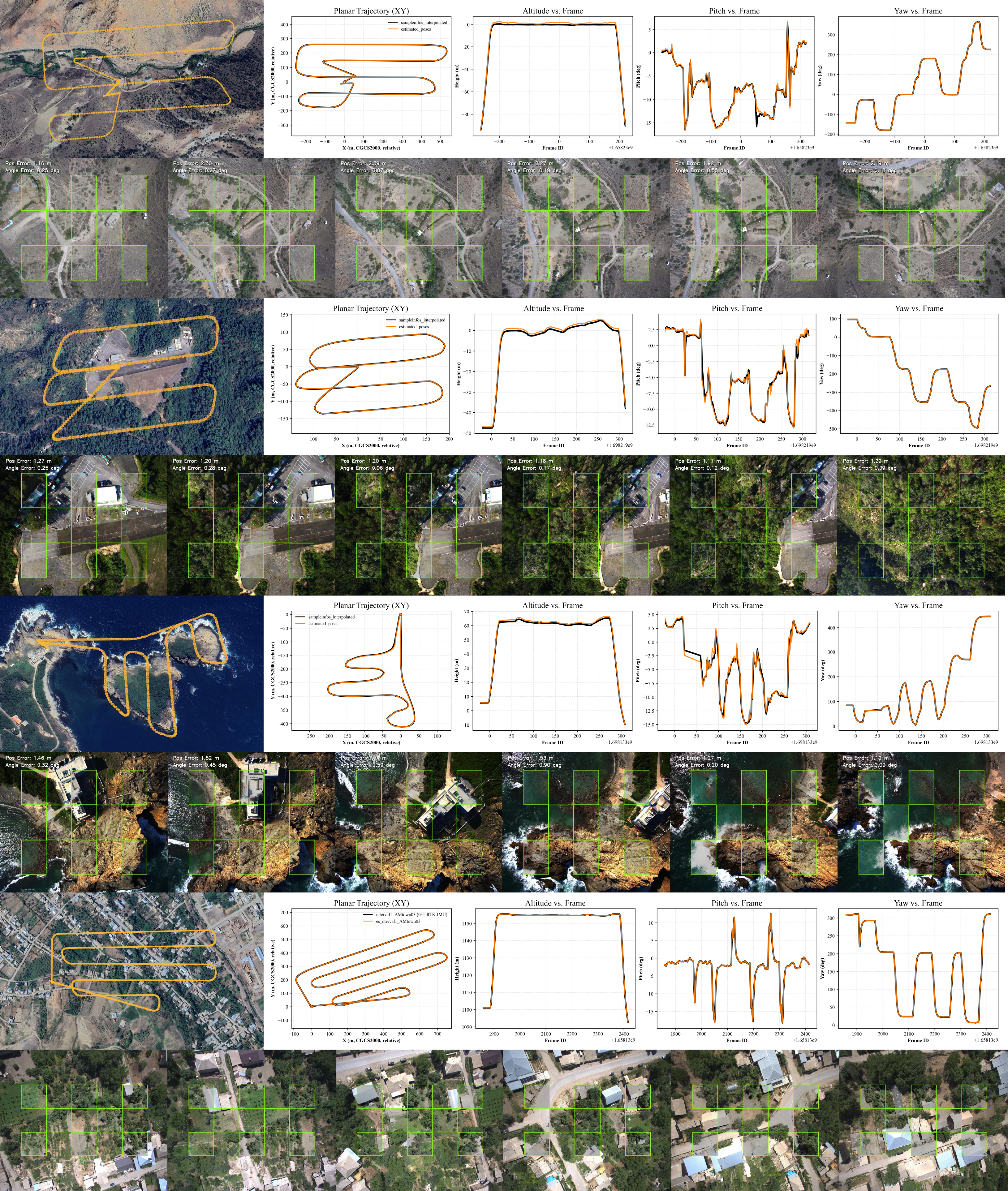}
  \caption{\textbf{Generalization performance on the standard \textit{UAVScenes} benchmark dataset.} This figure validates our model's strong generalization capability on four diverse, unseen scenes from the public UAVScenes benchmark. The scenes cover a wide range of environments: a town (\textit{AMtown}), a natural valley (\textit{AMvalley}), an airport (\textit{HKairport}), and an island (\textit{HKisland}).}
  \label{fig:uavscene_6frames}
\end{figure*}

\begin{table*}[t]
\centering
\caption{
\textbf{Summary of SynthCity-6 dataset.} Each trajectory provides camera frames with synchronized 6-DoF poses under varying \textit{weather, illumination, and altitude} conditions.
}
\scriptsize
\setlength{\tabcolsep}{3.5pt}
\begin{tabular}{lccccccc}
\toprule
\textbf{Sequence} & 
\textbf{Weather / Illumination} &
\textbf{Altitude (m)} & 
\textbf{Per Trajectory} &  
\textbf{Total Camera Frames} & 
\textbf{Location} \\
\midrule
Switzerland-seq4 & Foggy/Night/Rainy/Sunny/Sunset & 200/500 & 900 & 9000 & Thun, Switzerland \\
Switzerland-seq7 & Foggy/Night/Rainy/Sunny/Sunset & 200/500 & 900 & 9000 & Kirchlindach, Switzerland \\
Switzerland-seq12 & Foggy/Night/Rainy/Sunny/Sunset & 200/500 & 900 & 9000 & Lausanne, Switzerland \\
USA-seq2 & Foggy/Night/Cloudy/Sunny/Sunset & 200/500 & 900 & 9000 & Chicago, USA \\
USA-seq5 & Foggy/Night/Cloudy/Sunny/Sunset & 200/500 & 900 & 9000 & New York, USA \\
USA-seq8 & Foggy/Night/Cloudy/Sunny/Sunset & 200/500 & 900 & 9000 & California, USA \\
\midrule
Total & -- & -- & -- & 54k & -- \\
\bottomrule
\end{tabular}
\label{tab:SynthCity-6_summary}
\end{table*}

\begin{table*}[t]
\centering
\caption{
\textbf{Summary of UAVD4L-2yr dataset.} Each sequence contains camera frames captured under varying \textit{illumination and altitude}, covering different \textit{scene types} with annotated primary \textit{target objects}.
}
\scriptsize
\setlength{\tabcolsep}{3.5pt}
\begin{tabular}{lccccccc}
\toprule
\textbf{Sequence} & 
\textbf{Illumination} &
\textbf{Altitude (m)} &  
\textbf{Camera Frames} & 
\textbf{Scene Type \& Environment} &
\textbf{Target Object} \\
\midrule
UAVD4L-2yr-seq1 & Daytime & 70 & 900 & suburban & dynamic person\\
UAVD4L-2yr-seq2 & Daytime & 70 & 900 & urban & dynamic vehicles \\
UAVD4L-2yr-seq3 & Daytime & 70 & 900 & urban & static landmark \\
UAVD4L-2yr-seq4 & Daytime & 110 & 900 & suburban & dynamic vehicles \\
UAVD4L-2yr-seq5 & Night & 220 & 900 & urban & dynamic vehicles \\
UAVD4L-2yr-seq6 & Night & 310 & 900 & urban & dynamic vehicles \\
UAVD4L-2yr-seq7 & Night & 150 & 900 & suburban & dynamic vehicles \\
UAVD4L-2yr-seq8 & Night & 150 & 900 & urban & dynamic vehicles \\
\midrule
Total & -- & -- & 7.2k & -- & -- \\
\bottomrule
\end{tabular}
\label{tab:UAVD4L-2yr_summary}
\end{table*}

\begin{table*}[t]
\centering
\caption{
\textbf{Summary of UAVD4L-SynTarget dataset.} Each sequence contains camera frames captured under varying \textit{weather/illumination} conditions and \textit{flight altitudes}, with multiple \textit{target objects} annotated along with their quantities.
}
\scriptsize
\setlength{\tabcolsep}{3.5pt}
\begin{tabular}{lccccccc}
\toprule
\textbf{Sequence} & 
\textbf{Weather / Illumination} &
\textbf{Altitude (m)} &  
\textbf{Camera Frames} & 
\textbf{Target Object Number} \\
\midrule
UAVD4L-SynTarget-seq1 & Sunny & 130 & 304 & $>$100\\
UAVD4L-SynTarget-seq2 & Sunny & 130 & 344 & $>$100 \\
UAVD4L-SynTarget-seq3 & Sunny & 180 & 1136 & $>$100 \\
UAVD4L-SynTarget-seq4 & Foggy & 180 & 1136 & $>$100 \\
UAVD4L-SynTarget-seq5 & Night & 180 & 1136 & $>$100 \\
UAVD4L-SynTarget-seq6 & Foggy & 150 & 1192 & $>$100 \\
UAVD4L-SynTarget-seq7 & Sunny & 240 & 808 & $>$100 \\
\midrule
Total & -- & -- & 6k & -- \\
\bottomrule
\end{tabular}
\label{tab:UAVD4L-SynTarget_summary}
\end{table*}

\end{document}